\documentclass[letterpaper]{article} 
\usepackage{aaai2026}  
\usepackage{times}  
\usepackage{helvet}  
\usepackage{courier}  
\usepackage[hyphens]{url}  
\usepackage{graphicx} 
\usepackage{natbib}  
\usepackage{caption} 
\usepackage{algorithm}
\usepackage{algorithmic}

\usepackage{newfloat}
\usepackage{listings}
\DeclareCaptionStyle{ruled}{labelfont=normalfont,labelsep=colon,strut=off} 
\floatstyle{ruled}
\newfloat{listing}{tb}{lst}{}
\floatname{listing}{Listing}
\usepackage{algorithm}
\usepackage{algorithmic}

\usepackage{amsmath}
\usepackage{amsfonts}
\usepackage{multirow}
\usepackage{fontawesome5}
\usepackage{booktabs}
\usepackage{listings}
\usepackage{graphicx}
\usepackage[utf8]{inputenc} %
\usepackage[T1]{fontenc}    %
\usepackage{url}            %
\usepackage{booktabs}       %
\usepackage{amsfonts}       %
\usepackage{nicefrac}       %
\usepackage{microtype}      %
\usepackage{xcolor}         %
\usepackage{xspace}
\usepackage{mathtools}
\usepackage{listings}
\usepackage{makecell}
\usepackage{inconsolata}
\usepackage{textcomp}
\usepackage{adjustbox}
\usepackage{caption}
\usepackage{subcaption}
\usepackage{enumitem}
\usepackage{amssymb}
\usepackage{amsmath}  %
\usepackage{xcolor}   %
\usepackage{ bbold }
\usepackage{multirow}
\usepackage{xspace}

\usepackage[skins,breakable]{tcolorbox}

\usepackage{tikz}
\usetikzlibrary{tikzmark,decorations.pathreplacing,calc}

\newcommand\zll[1]{} %

\definecolor{codegreen}{rgb}{0,0.6,0}
\definecolor{codegray}{rgb}{0.5,0.5,0.5}
\definecolor{codepurple}{rgb}{0.58,0,0.82}
\definecolor{backcolour}{rgb}{0.95,0.95,0.92}

\lstdefinestyle{mystyle}{
    backgroundcolor=\color{backcolour},   
    commentstyle=\color{codegreen},
    keywordstyle=\color{magenta},
    numberstyle=\tiny\color{codegray},
    stringstyle=\color{codepurple},
    basicstyle=\ttfamily\scriptsize,
    breakatwhitespace=false,         
    breaklines=true,                 
    captionpos=b,                    
    keepspaces=true,                 
    numbers=left,                    
    numbersep=5pt,                  
    showspaces=false,                
    showstringspaces=false,
    showtabs=false,                  
    tabsize=2,
    escapeinside={(*@}{@*)}, 
    postbreak=\mbox{\textcolor{red}{$\hookrightarrow$}\space},
}

\lstnewenvironment{queryl}[1][] 
   {\lstset{frame=shadowbox,escapechar=`,linewidth=8cm, #1}}
   {}

\definecolor{light-purple}{RGB}{151,156,171}
\definecolor{blue-color}{RGB}{40,166,189}
\definecolor{pink-color}{RGB}{237,46,104} 
\definecolor{dark-grey-color}{RGB}{79,91,102}

\newcommand{\promptsubsection}[1]{
\setlength{\parskip}{6pt} \noindent\textbf{{#1}:}
}

\newtcolorbox[list inside=prompt,auto counter,number within=section]{prompt}[1][]{
    colbacktitle=black!80,
    colframe=black!80,
    coltitle=white,
    fontupper=\footnotesize,
    boxsep=5pt,
    left=0pt,
    right=0pt,
    top=0pt,
    bottom=0pt,
    boxrule=1pt,
    enhanced, 
    breakable,
    skin first=enhanced,
    skin middle=enhanced,
    skin last=enhanced,
    #1,
}

\title{ReCode: Updating Code API Knowledge with Reinforcement Learning}
\author{
Haoze Wu\textsuperscript{\rm 1},
Yunzhi Yao\textsuperscript{\rm 1},
Wenhao Yu\textsuperscript{\rm 2},
Ningyu Zhang\textsuperscript{\rm 1, 3}\thanks{Corresponding author.}
}

\affiliations{
\textsuperscript{\rm 1}Zhejiang University\\
\textsuperscript{\rm 2}Tencent AI, Seattle Lab\\
\textsuperscript{\rm 3}State Key Lab. for Novel Software Technology, Nanjing University, P.R. China\\[4pt]
wuhz1020@gmail.com, yyztodd@zju.edu.cn, wenhaoyu97@gmail.com, zhangningyu@zju.edu.cn
}

\usepackage{bibentry}

\begin{document}
\maketitle
\begin{abstract}
Large Language Models (LLMs) exhibit remarkable code generation capabilities but falter when adapting to frequent updates in external library APIs. This critical limitation, stemming from reliance on outdated API knowledge from their training data, even with access to current documentation, impedes reliable code generation in dynamic environments. To tackle this issue, we propose ReCode (rule-based \textbf{Re}inforcement learning for \textbf{Code} Update), a novel framework that mimics human programmer adaptation to API changes. Specifically, we construct a dataset of approximately 2,000 data entries to train the LLMs to perform version migration based on updated information. Then, we introduce a modified string similarity metric for code evaluation as the reward for reinforcement learning. Our experiments demonstrate that ReCode substantially boosts LLMs' code generation performance in dynamic API scenarios, especially on the unseen CodeUpdateArena task. Crucially, compared to supervised fine-tuning, ReCode has less impact on LLMs' general code generation abilities. We apply ReCode on various LLMs and reinforcement learning algorithms (GRPO and DAPO), all achieving consistent improvements. Notably, after training, Qwen2.5-Coder-7B outperforms that of the 32B parameter code instruction-tuned model and the reasoning model with the same architecture. 
\end{abstract}

\begin{links}
    \link{Code}{https://github.com/zjunlp/ReCode}
\end{links}

\section{Introduction}
\label{sec:intro}

\begin{figure}[!ht]
    \centering
    \includegraphics[width=\linewidth]{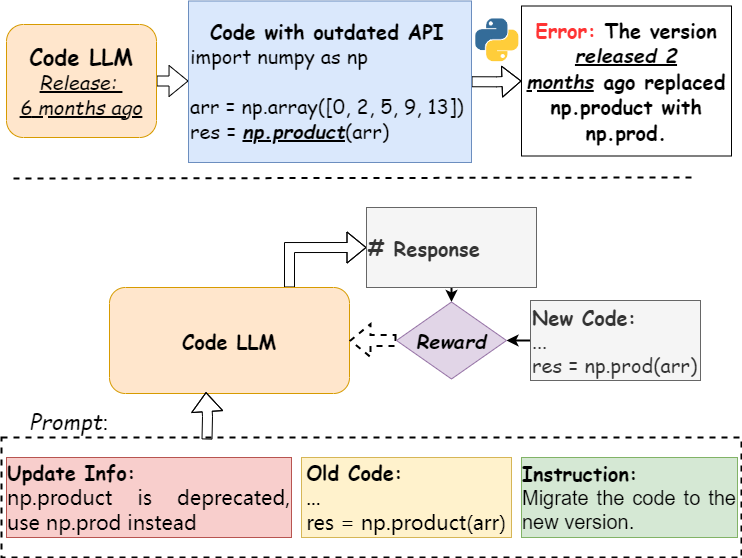}
    \caption{\textbf{Top}: LLMs cannot be aware of API updates that occur after their release date, which may lead to code errors. \textbf{Bottom}: Simply incorporating update information into the prompt cannot effectively alleviate the issue of outdated APIs. ReCode enhances their ability to migrate code to new versions through rule-based RFT.}
    \label{fig:overview}
\end{figure}

\begin{figure*}
    \centering
    \includegraphics[width=\linewidth]{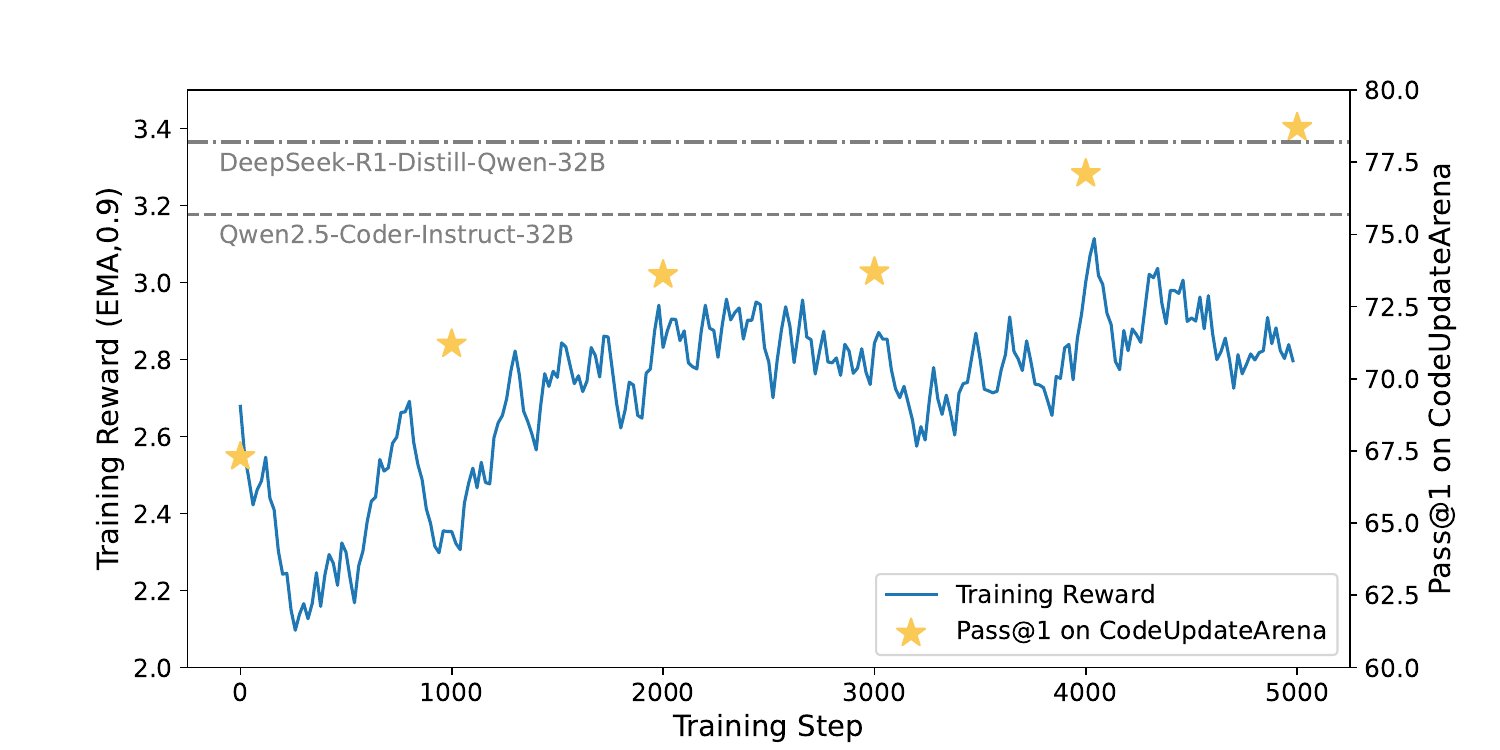}
    \caption{\textbf{Training Reward and Test Pass Rate during RL Fine-Tuning}. It demonstrates that Qwen2.5-Coder-7B-Instruct can enhance performance on the unseen CodeUpdateArena, even surpassing 32B code model and reasoning model after training. The two dashed lines in the figure represent the Pass@1 of the corresponding models on CodeUpdateArena.}
    \label{fig:head}
\end{figure*}

Large Language Models (LLMs) have recently demonstrated remarkable code generation abilities~\cite{chen2021evaluatinglargelanguagemodels,zan2023largelanguagemodelsmeet,rozière2024codellamaopenfoundation,deepseek-coder,hui2024qwen2,codegemmateam2024codegemmaopencodemodels,openai2024gpt4technicalreport,jiang2024surveylargelanguagemodels,geminiteam2025geminifamilyhighlycapable,qwen3}. This capacity enables LLMs to solve data science-related tasks by calling external libraries in the generated code~\cite{lai2022ds1000naturalreliablebenchmark,wang2025largelanguagemodelsreplace,hong2025data}. However, the APIs of external libraries are updated very frequently, while the model still has outdated information in their parameters~\cite{wu2024versicodeversioncontrollablecodegeneration,islah2024gitchameleonunmaskingversionswitchingcapabilities,wang2025llmsmeetlibraryevolution,yao2023editing,liu2025codeupdatearenabenchmarkingknowledgeediting}. As shown in Figure~\ref{fig:overview}-top, the LLM released before would generate code containing outdated APIs, leading to task failure in the up-to-date environment. This poses a challenge for the application of LLMs in scenarios that require high-quality code generation, such as AI4Research~\cite{lu2024aiscientist,starace2025paperbenchevaluatingaisability,tongyideepresearchteam2025tongyideepresearchtechnicalreport,liu2025webexplorerexploreevolvetraining}, Software Engineering~\cite{zhang2023survey,jimenez2024swebenchlanguagemodelsresolve,deng2025swebenchproaiagents}, and Human-Computer Interaction~\cite{zhang2024unifying,li2025tooldecathlonbenchmarkinglanguage}.

The root cause of these challenges lies in the fact that LLMs are trained on static datasets, making it difficult for them to adapt to dynamically evolving scenarios such as API updates. While supervised fine-tuning (SFT) can partially alleviate this issue~\cite{liu2025codeupdatearenabenchmarkingknowledgeediting}, the high frequency of updates may result in exorbitant costs and catastrophic forgetting of previously learned knowledge~\cite{biderman2024loralearnsforgets,luo2025empiricalstudycatastrophicforgetting}. An alternative approach involves embedding updated information directly into the prompt as a dynamic knowledge supplement~\cite{liu2025codeupdatearenabenchmarkingknowledgeediting,wang2025llmsmeetlibraryevolution}. This method can be further enhanced by integrating it with retrieval-augmented generation (RAG), offering greater potential in handling dynamic scenarios~\cite{lewis2021retrievalaugmentedgenerationknowledgeintensivenlp,gao2024retrievalaugmentedgenerationlargelanguage,gupta2024comprehensivesurveyretrievalaugmentedgeneration}. Furthermore, inputting updated knowledge through prompts avoids the risk of erasing other unrelated knowledge. Nevertheless, significant room for improvement remains in version-related code evaluation benchmarks, particularly for open-source code models~\cite{liu2025codeupdatearenabenchmarkingknowledgeediting}.
This may be attributed to the conflict between inherent (parameters) and external (prompt) knowledge, where LLMs tend to prioritize their internal knowledge.
In summary, while prompting is better suited for this scenario than SFT, it is necessary to address the challenge of knowledge conflicts.

Reinforcement Fine-Tuning (RFT) has been shown to enhance the model's ability to integrate retrieval in RAG systems, boosting response accuracy~\cite{jin2025searchr1trainingllmsreason}.
Inspired by this, we design ReCode, a rule-based \textbf{Re}inforcement learning for \textbf{Code} Update approach ~\cite{kaelbling1996reinforcementlearningsurvey,ghasemi2025comprehensivesurveyreinforcementlearning} to enhance the model's code migration capabilities when encountering conflicting information.
To take data science as an example, programmers first learn a specific version of the library, such as NumPy. Then, after being informed of the API updates, they can map the old version of the code in their minds to the new version. Our goal is to enable LLMs to use new APIs to complete tasks based on the updated information provided in the prompt. Similar to the Version-Aware Code Migration task in Versicode~\cite{wu2024versicodeversioncontrollablecodegeneration}, we train the model to migrate code from an old version to a new version based on the update information.
As shown in Figure~\ref{fig:overview}-bottom, we put the updated document in the prompt and fine-tune the model to better understand the prompt by using reinforcement learning based on the improved string similarity reward for code evaluation.

We evaluate the model's performance on CodeUpdateArena~\cite{liu2025codeupdatearenabenchmarkingknowledgeediting}, a more challenging task, where the model must solve practical tasks using API update information. Our experimental results indicate that ReCode significantly enhances LLMs' code generation capabilities in dynamic API scenarios. As shown in Figure~\ref{fig:head}, \textbf{the trained Qwen2.5-Coder-7B model outperforms Qwen2.5-Coder-32B and achieves a higher Pass@1 score than DeepSeek-R1-Distill-Qwen-32B}. Additionally, ReCode's impact on the model's general code generation abilities is less pronounced than that of supervised fine-tuning (SFT).
In summary, our contributions are: (1) ReCode is the first to explore the application of rule-based RFT in dynamic API scenarios. Particularly, we combine the use of prompts with RFT; (2) We construct a training dataset comprising approximately 2,000 data entries, specifically designed to train models in performing version migration based on updated information; (3) Through extensive experiments and analyses, we highlight the potential of ReCode in code generation and knowledge update scenarios.

\begin{figure*}[!ht]
    \centering
    \includegraphics[width=\linewidth]{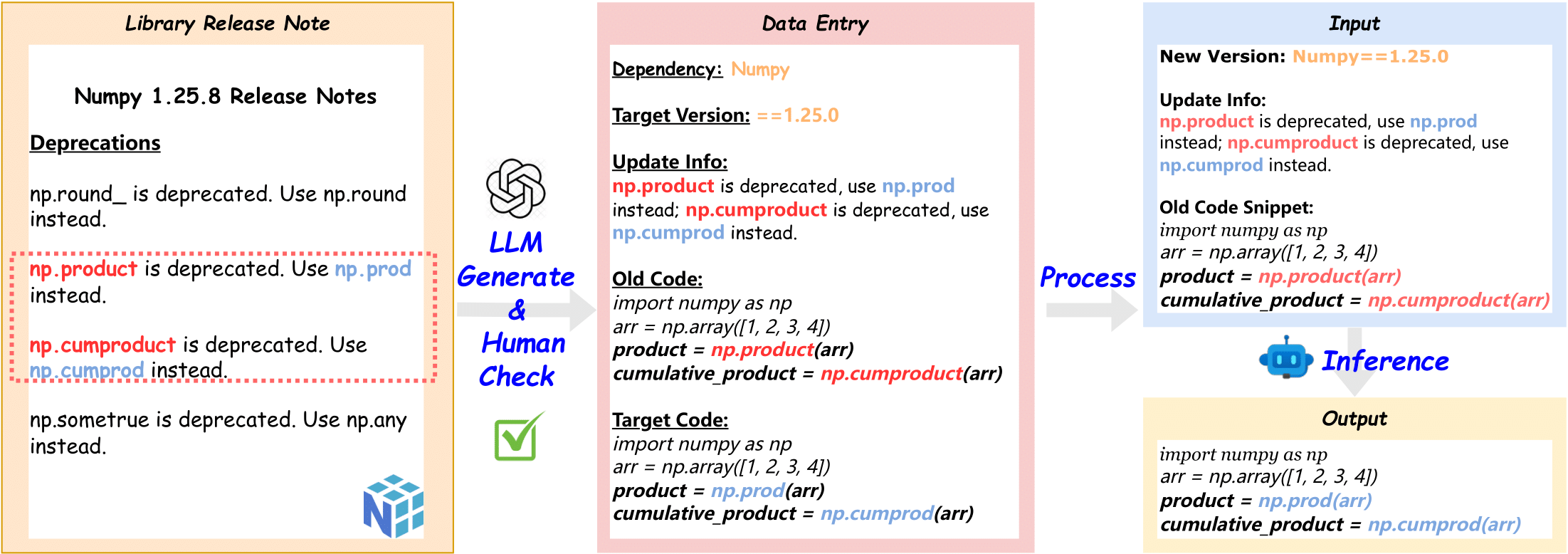}
    \caption{The pipeline of data collection and training task with a running example.}
    \label{fig:data_collect}
\end{figure*}

\section{API Knowledge Update}
\label{sec:api-kn-up}

\textbf{Problem formula}.
During the pre-training phase, the LLM accumulates code knowledge within its parameter $\boldsymbol{\theta}$ via an autoregressive approach~\cite{yenduri2023generativepretrainedtransformercomprehensive}. 
However, $\boldsymbol{\theta}$ remains static post-pre-training and thus fails to incorporate subsequent API updates. Consequently, when tasked with question $q$, the model may produce code containing outdated APIs: $\text{\textit{code with outdated API}} \gets P_{\boldsymbol{\theta}}(\cdot | q)$.

One basic strategy is to embed updated API information within the prompt $c_{update}$.
The model then generates code based on both the question $q$ and the embedded update details: $\text{\textit{code}} \gets P_{\boldsymbol{\theta}}(\cdot | q, c)$.
Nevertheless, this method introduces potential conflicts between the model's internal knowledge (housed in $\boldsymbol{\theta}$) and external information (provided in the prompt $c$).
Such conflicts can lead to the model overlooking the updated API details in $c$, even when additional updates are supplied.
Unlike directly writing the updated knowledge into the model's parameter $\boldsymbol{\theta}$, we employ reinforcement learning to fine-tune $\boldsymbol{\theta} \xrightarrow{} \boldsymbol{\theta}'$, enabling the model to more effectively leverage the updated API information presented in the prompt. This refined approach allows the model to generate code that better aligns with current API standards: $\text{\textit{update code}} \gets P_{\boldsymbol{\theta}'}(\cdot | q, c)$.

\noindent \textbf{CodeUpdateArena}.
CodeUpdateArena~\cite{liu2025codeupdatearenabenchmarkingknowledgeediting} is designed to evaluate the ability of LLMs to handle API updates.
The dataset comprises 670 program synthesis tasks, covering updates to 54 functions across seven different Python packages.
Unlike other datasets, CodeUpdateArena is a synthetic dataset generated with LLM assistance.
It prevents overlap with the model's training data and allows for the assessment of LLMs' adaptability to completely new updates.
In addition, each entry in the dataset includes at least three test cases that can be directly executed for verification.

\section{ReCode}

\subsection{Overview}
\label{sec:overview}

As we described in Section~\ref{sec:intro}, providing API update information to LLMs via prompts is the most promising solution, as it is more aligned with the behavior of human programmers and can be seamlessly integrated with RAG.
However, current LLMs usually miss the information provided in the instructions.
Our goal is to enhance the model's ability to follow the updated information provided in the prompt.

\textbf{Training}.
For training, we train models to perform version migration based on updated information.
As shown in Figure~\ref{fig:data_collect}-right, given a data entry $e_i=[d_i,v_i,u_i,c_i^{(old)},c_i^{(target)}]$, which corresponds to [\textit{Dependency, Target Version, Update Info, Old Code, and Target Code}], the input of training task is $x_i = [d_i,v_i,u_i,c_i^{(old)}]$ and the output is $y=c_i^{(target)}$.
The actual prompts with templates we used are provided in the Appendix.
We provide a detailed discussion of the training dataset construction in Section~\ref{sec:data}.
Moreover, we detail the rewards utilized in the training process in Section~\ref{sec:reward-design}.

\textbf{Testing}.
During testing, each piece of data in CodeUpdateArena~\cite{liu2025codeupdatearenabenchmarkingknowledgeediting} includes: Dependency and API update doc $u_i$, a real-world question $q_i$, and a function signature to be implemented $s_i$.
The test task requires the model to generate the complete function code for $s_i$ based on the above three pieces of information.
Since the dataset provides test cases, we use Pass@$k$ as the evaluation metric.
The prompts with templates are provided in the Appendix.

\begin{figure*}
    \centering
    \includegraphics[width=\linewidth]{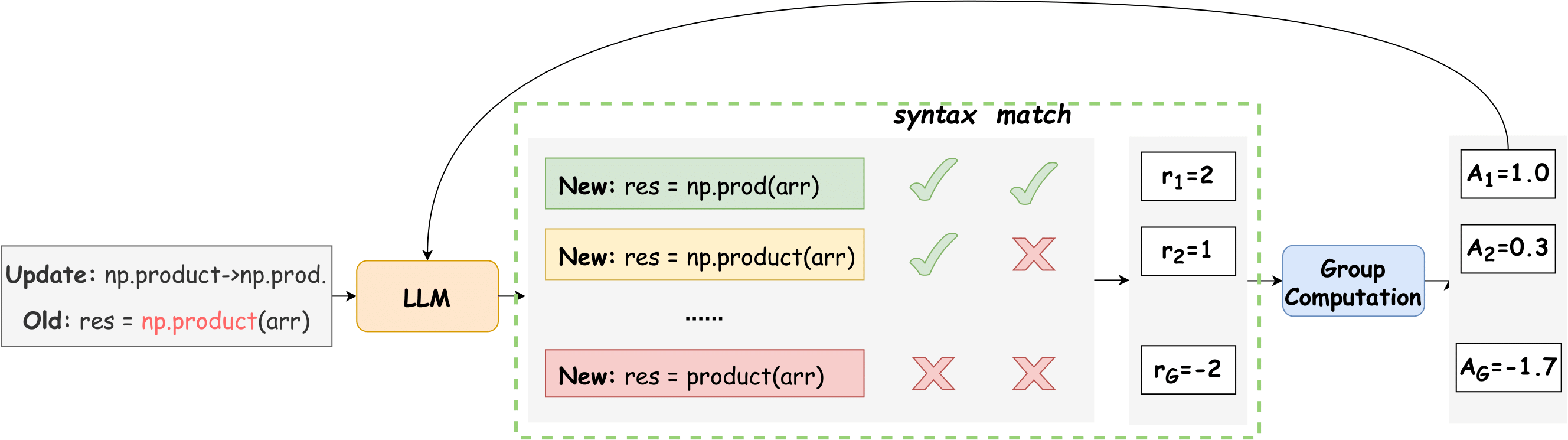}
    \caption{\textbf{Correctness Reward and Training Pipeline (taking GRPO as an example)}. The dashed box shows the correctness reward of our design, which includes two parts: syntax checking and string matching. It is worth mentioning that ReCode can be adapted to any reinforcement learning algorithm and is not limited to GRPO.}
    \label{fig:rft}
\end{figure*}

\subsection{Training Dataset Construction}
\label{sec:data}

Since no existing dataset provides input-output pairs, we construct our own training dataset.
As shown in Figure~\ref{fig:data_collect}-left, the data construction process is as follows:
\begin{enumerate}
    \item We access the release notes of major data science libraries (e.g., Numpy, Pandas, PyTorch, matplotlib) to identify paragraphs that detail specific API updates.
    \item We leverage GPT-4 to generate two code snippets with equivalent functionality: one using the old API and the other utilizing the updated API.
    \item Human experts review these code snippets to ensure they incorporate the updated API correctly. More details can be seen in Appendix.
    \item Only those code snippets that pass the expert review are included in the dataset.
\end{enumerate}

The final dataset comprises approximately 2K entries, with detailed statistics provided in Appendix. 
It is worth noting that the API updates in our dataset encompass a diverse range of changes, including but not limited to API renaming, parameter addition, and functionality modification. 
A detailed discussion is included in Appendix.

\textbf{Isolation of training and testing data}.
In addition to the different tasks of training and testing, there are also fundamental differences in the sources of their API updates.
The data we collected all come from real API updates; whereas CodeUpdateArena~\cite{liu2025codeupdatearenabenchmarkingknowledgeediting} is a completely synthetic dataset generated by LLMs. We believe that this fundamentally ensures the isolation between datasets.

\subsection{Reward Design}
\label{sec:reward-design}

Similar to the existing works~\cite{deepseekai2025deepseekr1incentivizingreasoningcapability,xie2025logicrlunleashingllmreasoning}, the reward in ReCode consists of two parts: format and correctness.

\textbf{Format Reward}.
We hope that the output of our model meets the format: 

\texttt{<think>...</think><answer>...</answer>}.

\noindent That is, to output the thinking process within the \texttt{<think>} tag, and to output the target code within the \texttt{<answer>} tag.
The format reward is defined as follows:
\begin{equation}
\begin{split}
    R_{format}(x)=\left\{
    \begin{matrix}
+1,& \text{if x meets the format}  \\
-1,& \text{else} 
\end{matrix}\right.
\end{split}
\end{equation}

\noindent where $x$ is the output of the model.

\textbf{Correctness Reward}.
Code, unlike math~\cite{hendrycks2021measuringmathematicalproblemsolving,liu2024mathbenchevaluatingtheoryapplication,chernyshev2025umath}, does not have a verifiable, unique standard answer. When improving coding skills, code quality can be verified by the pass rate of test cases~\cite{code-r1}. However, we argue that the pass rate of test cases is not a suitable reward metric for code migration task. Our training objective is to migrate correctly generated code to a new version, with the focus on ``migration'' rather than the inherent correctness of the code itself. 
Following previous works~\cite{wu2024versicodeversioncontrollablecodegeneration,wang2025llmsmeetlibraryevolution}, we use string matching metrics to evaluate the code's cross-version migration capability:
\begin{itemize}
    \item \textbf{Edit Similarity (ES)}: ES assesses the similarity between predicted completions and target codes by analyzing the edit operations needed to transform one into the other.
    \item \textbf{Exact Match (EM)}: EM calculates the rate at which predicted completions exactly match the target codes after normalizing return values.
\end{itemize}
Versicode~\cite{wu2024versicodeversioncontrollablecodegeneration} introduced the Critical Diff Check (CDC) metric by adding rules based on EM. However, since it only considered API name updates, CDC is not applicable to diverse API updates. Drawing on Logic-RL's~\cite{xie2025logicrlunleashingllmreasoning} reward mechanisms, we incorporated Abstract Syntax Tree (AST)-based code syntax checking into the string matching metrics, as illustrated in Figure~\ref{fig:rft}.
Taking EM as an example:
\begin{equation}
\begin{split}
    {EM^*}(x)=\left\{
    \begin{matrix}
+2.0,& \text{if x match the target}  \\
-1.5,& \text{elif x is syntactically valid} \\
-2.0,& \text{else}
\end{matrix}\right.
\end{split}
\end{equation}
Similarly, $ES^*$ is:
\begin{equation}
\begin{split}
    {ES^*}(x)=\left\{
    \begin{matrix}
-2.0, \text{if x isn't syntactically valid}  \\
ES(x)\times3.5-1.5, \text{else}
\end{matrix}\right.
\end{split}
\end{equation}
The reward range when the syntax is correct is also $[-1.5, 2.0]$.
Our reward is applicable to any RL algorithm that uses policy gradient updates. 
In our experiments, we select GRPO~\cite{shao2024deepseekmathpushinglimitsmathematical} and its modified version, DAPO~\cite{yu2025dapoopensourcellmreinforcement}, as the training algorithms.

\section{Experiment}

\begin{table*}[!ht]
    \centering
    \begin{tabular}{c|c|cc|c}
    \toprule
        \multirow{2}{*}{\textbf{Model}} &  \multirow{2}{*}{\textbf{Method}}    
        & \multicolumn{2}{c|}{\textbf{CodeUpdateArena}}
        & \multirow{2}{*}{\textbf{HumanEval+($\Delta$)}} \\ 
        \cmidrule(lr){3-4}  
              &                   
              & \textbf{Pass@\,$1$($\Delta$)} 
              & \textbf{Pass@\,$5$($\Delta$)} 
              &              
        \\ 
    \midrule
        Qwen2.5-Coder-32B-Instruct & Untrained & 75.7 \small{{(+0.0)}} & 84.3 \small{{(+0.0)}} & - \\
        DeepSeek-R1-Distill-Qwen-32B & Untrained & \textbf{78.2} \small{{(+0.0)}} & \textbf{86.1} \small{{(+0.0)}} & - \\
        \midrule
        \multirow{4}*{DS-v1.5-Coder-7B-Instruct} & Untrained & 59.1 \small{{(+0.0)}} & 72.5 \small{{(+0.0)}} & 71.3 \small{{(+0.0)}} \\
        ~ & SFT & 53.4 \small{{(-5.7)}} & 67.3 \small{{(-5.3)}} & 64.0 \small{{(-7.3)}} \\
        ~ & ReCode \small{GRPO} & 63.6 \small{{(+4.5)}} & 77.3 \small{{(+4.7)}} & 67.7 \small{{(-3.6)}}\\
        ~ & ReCode \small{DAPO} & 63.6 \small{{(+4.5)}} & 78.2 \small{{(+5.6)}} & 68.9 \small{{(-2.4)}} \\
        \midrule
        \multirow{4}*{Qwen2.5-Coder-7B-Instruct} & Untrained  & 67.3 \small{{(+0.0)}} & 74.0 \small{{(+0.0)}} & 84.1 \small{{(+0.0)}} \\
        ~ & SFT & 69.4 \small{{(+2.1)}} & 78.2 \small{{(+4.1)}} & 70.2 \small{{(-11.7)}} \\
        ~ & ReCode \small{GRPO} & 74.6 \small{{(+7.4)}} & 82.1 \small{{(+8.0)}} & 82.3 \small{{(-1.8)}} \\
        ~ & ReCode \small{DAPO} & \textbf{78.7} \small{{(+11.3)}} & 84.3 \small{{(+10.2)}} & 81.7 \small{{(-2.4)}}\\
        \bottomrule
    \end{tabular}
    \caption{The performance results using the GRPO and DAPO algorithms on CodeUpdateArena and HumanEval+.}
    \label{tab:main_result}
\end{table*}

\subsection{Experimental Setup}

\textbf{Model}.
The community commonly opts for rule-based reinforcement learning training of models in the Qwen family~\cite{yu2025dapoopensourcellmreinforcement,code-r1,zeng2025simplerlzooinvestigatingtamingzero}.
In our experiments, we use two code models to evaluate our method. This is to verify that our method can bring improvements not only to the Qwen model. The models we used are Qwen-2.5-Coder-7B-Instruct \cite{hui2024qwen2} and DeepSeek-v1.5-Coder-7B-Instruct \cite{deepseek-coder}. Both models demonstrate superior performance among models with fewer than 10B parameters. Notably, instruction-tuned models are found to retain R1 characteristics \cite{xie2025logicrlunleashingllmreasoning}.
Additionally, in the experimental section, we employ Qwen2.5-Coder-Instruct-32B \cite{hui2024qwen2} and DeepSeek-R1-Distill-Qwen-32B \cite{deepseekai2025deepseekr1incentivizingreasoningcapability} as baseline models. 
The performance metrics for these two models are obtained via the API of SiliconFlow\footnote{\url{https://www.siliconflow.cn/}}.
A detailed discussion on baseline selection is left in Appendix.

\noindent \textbf{Training}.
In our experiment, we utilize the DoRA~\cite{liu2024doraweightdecomposedlowrankadaptation} to update the model due to the computational limits. 
The hyperparameters are configured as $r=64,\alpha=64$.
For the RL component, both GRPO and DAPO are set with $G=8$, while in GRPO, the parameter $\beta$ is assigned a value of $0.001$.
The training procedure spans 5000 steps, targeting the training tasks with a batch size set at $8$ and a learning rate of $5\times10^{-5}$.
Specifically, the initial $150$ training steps incorporate a learning rate warm-up schedule, which is subsequently followed by a cosine schedule.

\noindent \textbf{Test Dataset and Metric}.
We select the CodeUpdateArena benchmark~\cite{liu2025codeupdatearenabenchmarkingknowledgeediting} as our test task and filter inaccurate cases from this public dataset\footnote{\url{https://github.com/leo-liuzy/CodeUpdateArena/blob/main/data/arena-ungrouped.jsonl}}. 
A detailed account of these errors is provided in Appendix. 
We evaluate our corrected data using the \textbf{Pass@$k$} metric, with $k$ values of $1$ and $5$ in our experiments.
All tests are conducted using the inputs with updated information as described in Section~\ref{sec:overview}. This in itself represents the setting of RAG.
To examine how additional training affects the model's general code capabilities, we employed HumanEval+~\cite{liu2023is}.

\subsection{Main Result}

Table~\ref{tab:main_result} presents the results obtained from CodeUpdateArena after training.

\textbf{ReCode enhances the model's pass rate within the arena}. We conduct an ablation study on the reward components in Section~\ref{sec:ablation} and adopt ES* as the training reward.
As shown in Table~\ref{tab:main_result}, both GRPO and DAPO yield consistent gains on the arena leaderboard.
Most notably, the Qwen2.5-Coder-7B-Instruct model ultimately surpasses both the 32B-parameter instruction-tuned model and the distilled reasoning model of the same architecture on Pass@1.
Its Pass@5 is slightly lower than that of DeepSeek-R1-Distill-Qwen-32B, a finding that echoes prior work~\cite{yue2025doesreinforcementlearningreally}: the ceiling of reasoning improvements delivered by RL is capped by the base model’s intrinsic capacity.
Crucially, the benefits of ReCode are not confined to Qwen models: DeepSeek-Coder also enjoys a substantial uplift (Pass@1 +4.5, Pass@5 +5.6).
Considering that DeepSeek-Coder is inherently weaker than Qwen2.5-Coder, the observed disparity in performance gains remains entirely reasonable.

\textbf{SFT exhibits limited generalization capabilities when transitioning from the code migration task to real-world code generation tasks}. As Table~\ref{tab:main_result} shows, SFT’s performance gains lag behind those of ReCode (Qwen), and in some cases even degrade the pre-trained model’s efficacy (DS), consistent with observations by \citealp{liu2025codeupdatearenabenchmarkingknowledgeediting}. This limitation stems from SFT’s tendency to minimize loss by memorizing prompt–answer pairs while ignoring update documentation, leading to poor robustness under task discrepancy. In contrast, RL encourages the model to earn high rewards only by correctly interpreting documentation and generating valid code, thereby fostering genuine understanding and enabling consistent generalization across dynamic, decoupled training–testing settings.

\textbf{ReCode has less impact on the general capabilities of LLMs than SFT}.
Training the model on the updated API data may degrade its general-purpose coding proficiency.
To quantify this, we evaluate the model on HumanEval+~\cite{liu2023is}, a benchmark specifically designed to assess general code-generation abilities, after training.
As shown in Table~\ref{tab:main_result}, comparing pre-training and post-training results reveals that ReCode exerts a far milder influence on general coding capabilities than SFT.
Consequently, we maintain that ReCode remains the most viable and promising solution for dynamic API scenarios.

\subsection{Reward Design Ablation}
\label{sec:ablation}

In Section~\ref{sec:reward-design}, we outline the design space for the correctness reward, which encompasses EM, ES, EM*, and ES* metrics. Using Qwen2.5-Coder in conjunction with GRPO~\cite{shao2024deepseekmathpushinglimitsmathematical}, we conduct a comparative analysis to evaluate how these different reward mechanisms influence the final performance.

\begin{table}[]
    \centering
    \begin{tabular}{c|c|cc|cc}
        \toprule
         & format & +EM & +ES & +EM* & +ES* \\
         \midrule
        Pass@1($\Delta$) & -2.3 & -1.2 & +5.4 & +1.1 & +7.4 \\
        Pass@5($\Delta$) & -3.0 & -3.2 & +5.2 & +2.0 & +8.0 \\
         \bottomrule
    \end{tabular}
    \caption{Ablation study on reward design using Qwen2.5-Coder and GRPO, comparing the impact of format reward, exact match (EM), edit similarity (ES), and their syntax-checked variants (EM*, ES*) on final performance metrics (Pass@1 and Pass@5).}
    \label{tab:ablation}
\end{table}

Table~\ref{tab:ablation} illustrates the variations in the final metrics when training with different rewards. Below are three key findings:

\textbf{Using only the format reward does not lead to improvement}. Without correctness rewards, the policy model cannot receive feedback on whether the code migration is correct. Moreover, instruction-tuned models can follow instructions well without additional training; further training using only format rewards would be redundant. This underscores the necessity of introducing correctness rewards in the API update scenario.

\textbf{The inclusion of AST-based syntax checking is necessary}. It is evident that regardless of whether using the strict matching metric EM or the loose similarity measure ES, the inclusion of syntax checking enhances performance. We speculate that pure string matching may result in the model's understanding of the task degrading. Additionally, Versicode~\cite{wu2024versicodeversioncontrollablecodegeneration} also observed that when the target code length increases, the correlation coefficient between EM and the pass rate of test cases decreases.

\textbf{Similarity measurement is better than strict matching}. We find that although ES and EM ultimately share the same goal of aligning the output with the target code, ES and ES* achieve superior results. This may be related to the calculation of the inter-group advantage in naive GRPO. When none of the outputs in a group strictly match the target code, each output receives the same reward, leading to a zero advantage for each output. DAPO~\cite{yu2025dapoopensourcellmreinforcement} also highlighted this issue. In contrast, ES offers more flexible values. Even if none of the outputs in a group strictly match the target code, different outputs can still receive varying rewards.
To verify this, we calculate the proportion of data with inter-group reward variance being $0$ before training started.
To better highlight the differences, we discard the data with an initial reward of $2$.
As shown in Figure~\ref{fig:dynamic}-left, ES* has a lower proportion than EM*. 
This means the model can get reward feedback on more data it couldn't handle before (reward $< 2$) when using GRPO.

Unless otherwise specified, we employ ES* as the correctness reward in all other experiments.

\section{Analysis}

If not specifically emphasized otherwise, the analyses are based on Qwen2.5-Coder-7B-Instruct w/ DAPO ReCode.

\subsection{Training Dynamics}

\textbf{Reward and Pass Rate}.
As depicted in Figure~\ref{fig:head}, during the initial few hundred training steps, the reward undergoes a decline rather than an increase. This may appear counterintuitive, but our analysis reveals that the primary driver of this early-stage reduction is the decrease in format reward. Instruction-tuned models inherently possess robust instruction-following abilities, enabling them to readily comply with reasoning formats without undergoing training. Nevertheless, this very strength in adhering to instructions might concurrently restrict the exploration space of reinforcement learning algorithms, potentially entrenching specific modes of thinking. We posit that the initial reward drop reflects the model's reluctant shift as it commences exploration. It discards its pre-existing instruction-following capabilities and established thought processes to embark on a new learning phase. The subsequent reward increase is attributed to the model having sufficiently expanded its exploration space, which in turn enables the gradual enhancement of both reward and performance metrics. However, in our experimental observations, even with this initial reward decline, instruction-tuned models demonstrated a faster learning pace compared to their base model counterparts.

\begin{figure}
    \centering
    \includegraphics[width=\linewidth]{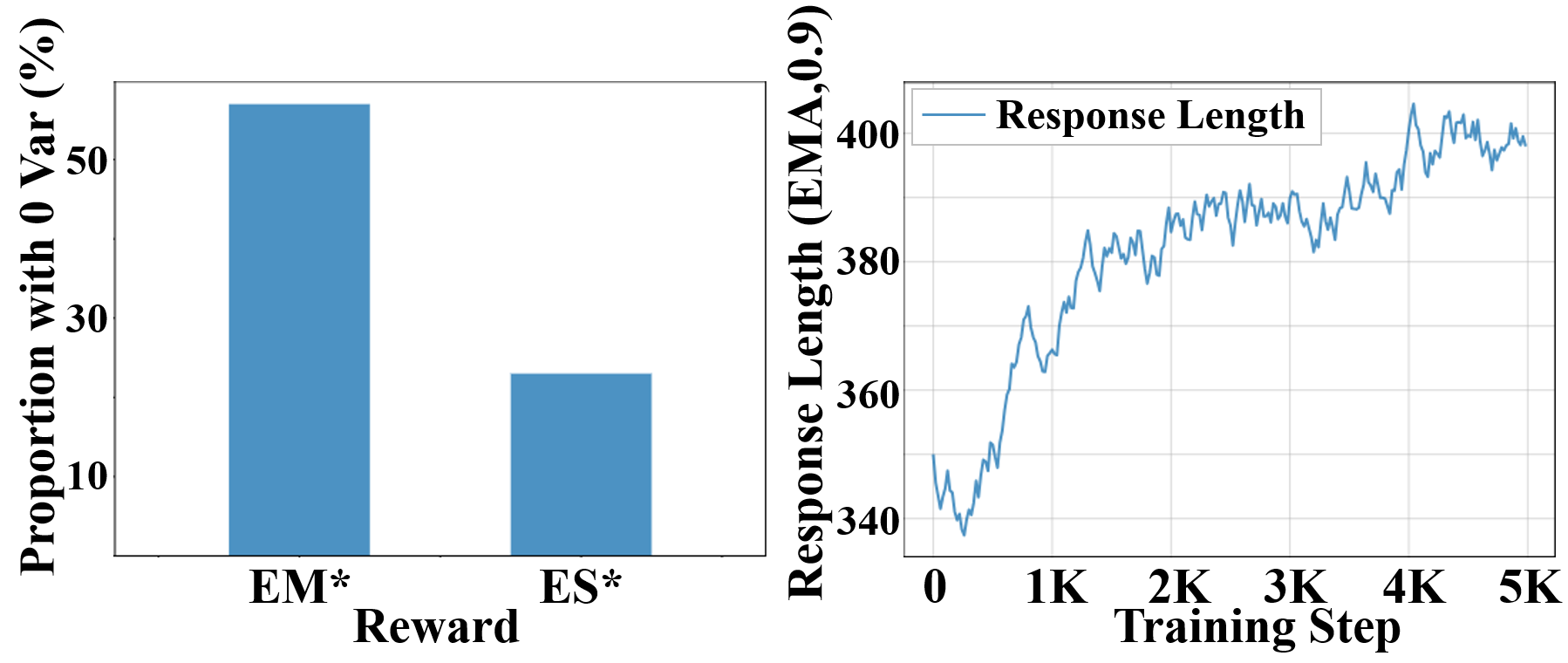}
    \caption{\textbf{Left}: the proportion of data with inter-group reward variance being $0$. \textbf{Right}: The changes in response length during the training process.}
    \label{fig:dynamic}
\end{figure}

\noindent\textbf{Response Length}.
As shown in Figure~\ref{fig:dynamic}-right, the model's output length tends to increase during training. However, we do not believe there is a direct relationship between the model's reasoning ability and response length. In our training tasks, the length of the model's Chain of Thought (CoT) output is inherently limited (around 300), and even after training, the model's output length only increases to approximately 400. This may be due to the limited thinking required for the migration task, with the model not engaging in excessive thinking throughout the process.

\subsection{Case Study: The Impact of RL?}

A natural question arises: Why does ReCode result in improvements? 
To explore this question, we divide the error cases before training into two categories. One category includes cases that are corrected after training, and the other includes those that remain wrong.

\textbf{ReCode can help LLMs overcome their laziness in acquiring external knowledge}.
We observe that the pretrained model exhibits "laziness" regarding the updates in the prompt. Specifically, it either ignores the mentioned updated APIs or overlooks the new API parameters. In the Appendix, we show an erroneous case where the model disregards the updated API \textit{math.sin} and misinterprets the angle unit "gradian," producing incorrect code.

\begin{tcolorbox}[colframe = black, colback = white]
\textbf{Hypothesis}. \textit{When generating code, an LLM draws from two information sources: internal knowledge (parameters) and provided prompts. Similar to humans, the model tends to rely more on its internal knowledge.}
\end{tcolorbox}

\noindent This propensity limits improvements when using prompts as the sole information source in dynamic scenarios.
This is also one of the challenges faced by RAG systems.
\citealt{longpre2022entitybasedknowledgeconflictsquestion} found that when encountering conflicting information between context and internal knowledge, QA models tend to rely on internal knowledge.
\citealt{xu2024knowledgeconflictsllmssurvey} emphasized that knowledge obsolescence can lead to a conflict between parameters and prompts.
Fortunately, ReCode corrects these issues, 
as demonstrated in the Appendix, 
where the model accurately uses the updated API information.
ReCode enables the LLM to overcome its tendency to be ``lazy'' in utilizing external knowledge.

\textbf{The capacity to address problems is fundamentally constrained by the pre-trained model}. 
Most uncorrected cases are unrelated to API updates. In the Appendix, we show a case where the model is supposed to develop a pendulum simulation program. It ignores the physical quantities in the parameters and fails to describe the physical laws as required in the prompt. Unsurprisingly, even after training, the model still couldn't write this function correctly under the condition of Pass@$1$. Our method doesn't focus on enhancing physical reasoning abilities. Thus, Qwen2.5-Coder-7B-Instruct, which originally lacked this ability, still couldn't complete the task after training.

\subsection{Evaluation Under Complex Scenarios}

Since our training and testing tasks only involve a single API update at a time, we explore two more complex scenarios:

\textbf{Multiple updates to a single API}. \textit{torch.gels()} was deprecated in version 1.2 in favor of \textit{torch.lstsq()}. Later, \textit{torch.lstsq()} was also deprecated in version 1.9, replaced by \textit{torch.linalg.lstsq()}. To update code from \textit{torch.gels} to \textit{torch.linalg.lstsq}, one could do it in two steps following the version sequence. However, as shown in Appendix, with update-aware prompting, the model can directly migrate from \textit{torch.gels} to \textit{torch.linalg.lstsq} without going through \textit{torch.lstsq}, which is more align with human programmers.

\textbf{Multiple API Updates}.
Can the model handle updates to multiple APIs at the same time? In the Appendix, we provide an example where the model correctly generated code when facing multiple API updates. It also doesn't show laziness in using the knowledge from the prompt.
To demonstrate the improvement achieved by ReCode, we construct 20 similar questions involving multiple API updates to test models' performance.
As shown in Table~\ref{tab:mul-api}, ReCode enhances the performance of Qwen2.5-Coder, bringing a 7B instruct model closer to the performance of a 32B reasoning model.

\begin{table}[]
    \centering
    \begin{tabular}{c|cc}
    \toprule
    Model  & Pass@1 & Pass@5 \\
    \midrule
      DeepSeek-R1-Distill-Qwen-32B & 65 & 80 \\
    \midrule
      Qwen2.5-Coder-7B-Instruct  & 35 & 55 \\
      +ReCode   & 60 & 75 \\
    \bottomrule
    \end{tabular}
    \caption{The results tested on 20 questions involving multiple API updates.}
    \label{tab:mul-api}
\end{table}

\section{Related Work}

\textbf{Benchmarking Code Generation}.
Most current code completion benchmarks \cite{chen2021codex,austin2021programsynthesislargelanguage,10103177,nijkamp2023codegenopenlargelanguage,zheng2024codegeexpretrainedmodelcode,zhuo2025bigcodebenchbenchmarkingcodegeneration} fail to test models' cross-version code migration skills.
To address this issue, several version-related code completion benchmarks have been proposed recently. Versicode \cite{wu2024versicodeversioncontrollablecodegeneration} includes over 9,000 code samples from websites but only covers API name changes and lacks executable test cases. In contrast, GitChameleon \cite{islah2024gitchameleonunmaskingversionswitchingcapabilities} collects over 100 real-world update samples to evaluate whether models can generate correct code based on library versions. CodeUpdateArena \cite{liu2025codeupdatearenabenchmarkingknowledgeediting} is a dataset of synthetic data with 670 samples generated by LLMs that includes executable test cases.

\noindent\textbf{Reinforcement Fine-Tuning}.
Reinforcement Learning (RL) remains crucial for LLM post-training, with RLHF being key to GPT's evolution into ChatGPT \cite{ouyang2022traininglanguagemodelsfollow}. With the emergence of slow-thinking reasoning models \cite{gpt-o1,deepseekai2025deepseekr1incentivizingreasoningcapability,kimiteam2025kimik15scalingreinforcement}, the community has recognized the importance of rule-based RL. 
Particularly, the "Aha moments" observed in mathematical tasks are quite fascinating.
Recently, the community has found that RFT is also effective in other domain tasks, such as video understanding~\cite{feng2025videor1reinforcingvideoreasoning,chen2025scalingrllongvideos}, code generation~\cite{code-r1,ma2025sqlr1trainingnaturallanguage}, tool using~\cite{li2025torlscalingtoolintegratedrl,qian2025toolrlrewardtoollearning,wang2025otcoptimaltoolcalls}, machine translation~\cite{feng2025mtr1zeroadvancingllmbasedmachine}, and others~\cite{jin2025searchr1trainingllmsreason}.

\section{Discussion and Conclusion}

To the best of our knowledge, ReCode is the first to adopt the rule-based RFT post-training method for dynamic API scenarios. By training models on the code migration task, we've addressed the issue of AI programmers struggling to migrate code based on updates without test data.
ReCode helps models make correct choices when encountering knowledge conflicts.
AI programmers are tireless and efficient but lack dynamic adaptability to new knowledge, a reasoning ability once thought unique to humans. How to endow AI with this ability remains a significant challenge.

ReCode represents an exploratory step in code generation. Through ReCode, Qwen2.5-Coder-7B-Instruct has achieved cross-scale performance, surpassing the strong reasoning model with 32B parameters in the Pass@1 metric.
In the future, we aim to explore this paradigm's application in other knowledge update scenarios and further enhance reward design and training algorithms.

\section*{Acknowledgement}
This work was supported by the National Natural Science Foundation of China (No. 62576307, No. NSFCU23B2055, No. NSFCU19B2027), the Fundamental Research Funds for the Central Universities (226-2023-00138), Ningbo Natural Science Foundation (2024J020), Yongjiang Talent Introduction Programme (2021A-156-G), Tencent AI Lab Rhino-Bird Focused Research Program (RBFR2024003), Information Technology Center and State Key Lab of CAD\&CG, Zhejiang University.

\small\bibliography{aaai2026}

\begin{thebibliography}{72}
\providecommand{\natexlab}[1]{#1}

\bibitem[{Austin et~al.(2021)Austin, Odena, Nye, Bosma, Michalewski, Dohan, Jiang, Cai, Terry, Le, and Sutton}]{austin2021programsynthesislargelanguage}
Austin, J.; Odena, A.; Nye, M.; Bosma, M.; Michalewski, H.; Dohan, D.; Jiang, E.; Cai, C.; Terry, M.; Le, Q.; and Sutton, C. 2021.
\newblock Program Synthesis with Large Language Models.
\newblock arXiv:2108.07732.

\bibitem[{Biderman et~al.(2024)Biderman, Portes, Ortiz, Paul et~al.}]{biderman2024loralearnsforgets}
Biderman, D.; Portes, J.; Ortiz, J. J.~G.; Paul, M.; et~al. 2024.
\newblock LoRA Learns Less and Forgets Less.
\newblock arXiv:2405.09673.

\bibitem[{Cassano et~al.(2023)Cassano, Gouwar, Nguyen, Nguyen, Phipps-Costin, Pinckney, Yee, Zi, Anderson, Feldman, Guha, Greenberg, and Jangda}]{10103177}
Cassano, F.; Gouwar, J.; Nguyen, D.; Nguyen, S.; Phipps-Costin, L.; Pinckney, D.; Yee, M.-H.; Zi, Y.; Anderson, C.~J.; Feldman, M.~Q.; Guha, A.; Greenberg, M.; and Jangda, A. 2023.
\newblock MultiPL-E: A Scalable and Polyglot Approach to Benchmarking Neural Code Generation.
\newblock \emph{IEEE Transactions on Software Engineering}, 49(7): 3675--3691.

\bibitem[{Chen et~al.(2021{\natexlab{a}})Chen, Tworek, Jun, Yuan et~al.}]{chen2021evaluatinglargelanguagemodels}
Chen, M.; Tworek, J.; Jun, H.; Yuan, Q.; et~al. 2021{\natexlab{a}}.
\newblock Evaluating Large Language Models Trained on Code.
\newblock arXiv:2107.03374.

\bibitem[{Chen et~al.(2021{\natexlab{b}})Chen, Tworek, Jun, Yuan et~al.}]{chen2021codex}
Chen, M.; Tworek, J.; Jun, H.; Yuan, Q.; et~al. 2021{\natexlab{b}}.
\newblock Evaluating Large Language Models Trained on Code.

\bibitem[{Chen et~al.(2025)Chen, Huang, Shi, Hu, Ye et~al.}]{chen2025scalingrllongvideos}
Chen, Y.; Huang, W.; Shi, B.; Hu, Q.; Ye, H.; et~al. 2025.
\newblock Scaling RL to Long Videos.
\newblock arXiv:2507.07966.

\bibitem[{Chernyshev et~al.(2025)Chernyshev, Polshkov, Artemova, Myasnikov, Stepanov, Miasnikov, and Tilga}]{chernyshev2025umath}
Chernyshev, K.; Polshkov, V.; Artemova, E.; Myasnikov, A.; Stepanov, V.; Miasnikov, A.; and Tilga, S. 2025.
\newblock U-{MATH}: A University-Level Benchmark for Evaluating Mathematical Skills in {LLM}s.

\bibitem[{Chu et~al.(2025)Chu, Zhai, Yang, Tong, Xie, Schuurmans, Le, Levine, and Ma}]{chu2025sftmemorizesrlgeneralizes}
Chu, T.; Zhai, Y.; Yang, J.; Tong, S.; Xie, S.; Schuurmans, D.; Le, Q.~V.; Levine, S.; and Ma, Y. 2025.
\newblock SFT Memorizes, RL Generalizes: A Comparative Study of Foundation Model Post-training.
\newblock arXiv:2501.17161.

\bibitem[{DeepSeek-AI(2025)}]{deepseekai2025deepseekr1incentivizingreasoningcapability}
DeepSeek-AI. 2025.
\newblock DeepSeek-R1: Incentivizing Reasoning Capability in LLMs via Reinforcement Learning.
\newblock arXiv:2501.12948.

\bibitem[{Deng et~al.(2025)Deng, Da, Pan, He, Ide, Garg et~al.}]{deng2025swebenchproaiagents}
Deng, X.; Da, J.; Pan, E.; He, Y.~Y.; Ide, C.; Garg, K.; et~al. 2025.
\newblock SWE-Bench Pro: Can AI Agents Solve Long-Horizon Software Engineering Tasks?
\newblock arXiv:2509.16941.

\bibitem[{Feng et~al.(2025{\natexlab{a}})Feng, Gong, Li, Guo, Wang, Peng, Wang, and Yue}]{feng2025videor1reinforcingvideoreasoning}
Feng, K.; Gong, K.; Li, B.; Guo, Z.; Wang, Y.; Peng, T.; Wang, B.; and Yue, X. 2025{\natexlab{a}}.
\newblock Video-R1: Reinforcing Video Reasoning in MLLMs.
\newblock arXiv:2503.21776.

\bibitem[{Feng et~al.(2025{\natexlab{b}})Feng, Cao, Ren, Su, Chen, Zhang, Xu, Hu, Wu, and Liu}]{feng2025mtr1zeroadvancingllmbasedmachine}
Feng, Z.; Cao, S.; Ren, J.; Su, J.; Chen, R.; Zhang, Y.; Xu, Z.; Hu, Y.; Wu, J.; and Liu, Z. 2025{\natexlab{b}}.
\newblock MT-R1-Zero: Advancing LLM-based Machine Translation via R1-Zero-like Reinforcement Learning.
\newblock arXiv:2504.10160.

\bibitem[{Gao et~al.(2024)Gao, Xiong, Gao, Jia, Pan, Bi, Dai, Sun, Wang, and Wang}]{gao2024retrievalaugmentedgenerationlargelanguage}
Gao, Y.; Xiong, Y.; Gao, X.; Jia, K.; Pan, J.; Bi, Y.; Dai, Y.; Sun, J.; Wang, M.; and Wang, H. 2024.
\newblock Retrieval-Augmented Generation for Large Language Models: A Survey.
\newblock arXiv:2312.10997.

\bibitem[{Ghasemi, Moosavi, and Ebrahimi(2025)}]{ghasemi2025comprehensivesurveyreinforcementlearning}
Ghasemi, M.; Moosavi, A.~H.; and Ebrahimi, D. 2025.
\newblock A Comprehensive Survey of Reinforcement Learning: From Algorithms to Practical Challenges.
\newblock arXiv:2411.18892.

\bibitem[{Guo et~al.(2024)Guo, Zhu, Yang, Xie, Dong, Zhang, Chen, Bi, Wu, Li, Luo, Xiong, and Liang}]{deepseek-coder}
Guo, D.; Zhu, Q.; Yang, D.; Xie, Z.; Dong, K.; Zhang, W.; Chen, G.; Bi, X.; Wu, Y.; Li, Y.; Luo, F.; Xiong, Y.; and Liang, W. 2024.
\newblock DeepSeek-Coder: When the Large Language Model Meets Programming -- The Rise of Code Intelligence.

\bibitem[{Gupta, Ranjan, and Singh(2024)}]{gupta2024comprehensivesurveyretrievalaugmentedgeneration}
Gupta, S.; Ranjan, R.; and Singh, S.~N. 2024.
\newblock A Comprehensive Survey of Retrieval-Augmented Generation (RAG): Evolution, Current Landscape and Future Directions.
\newblock arXiv:2410.12837.

\bibitem[{Hendrycks et~al.(2021)Hendrycks, Burns, Kadavath, Arora, Basart, Tang, Song, and Steinhardt}]{hendrycks2021measuringmathematicalproblemsolving}
Hendrycks, D.; Burns, C.; Kadavath, S.; Arora, A.; Basart, S.; Tang, E.; Song, D.; and Steinhardt, J. 2021.
\newblock Measuring Mathematical Problem Solving With the MATH Dataset.
\newblock arXiv:2103.03874.

\bibitem[{Hong et~al.(2025)Hong, Lin, Liu, Liu, Wu et~al.}]{hong2025data}
Hong, S.; Lin, Y.; Liu, B.; Liu, B.; Wu, B.; et~al. 2025.
\newblock Data Interpreter: An {LLM} Agent For Data Science.

\bibitem[{Hui et~al.(2024)Hui, Yang, Cui, Yang, Liu, Zhang et~al.}]{hui2024qwen2}
Hui, B.; Yang, J.; Cui, Z.; Yang, J.; Liu, D.; Zhang, L.; et~al. 2024.
\newblock Qwen2. 5-Coder Technical Report.
\newblock \emph{arXiv preprint arXiv:2409.12186}.

\bibitem[{Islah et~al.(2024)Islah, Gehring, Misra, Muller, Rish, Zhuo, and Caccia}]{islah2024gitchameleonunmaskingversionswitchingcapabilities}
Islah, N.; Gehring, J.; Misra, D.; Muller, E.; Rish, I.; Zhuo, T.~Y.; and Caccia, M. 2024.
\newblock GitChameleon: Unmasking the Version-Switching Capabilities of Code Generation Models.
\newblock arXiv:2411.05830.

\bibitem[{Jiang et~al.(2024)Jiang, Wang, Shen, Kim, and Kim}]{jiang2024surveylargelanguagemodels}
Jiang, J.; Wang, F.; Shen, J.; Kim, S.; and Kim, S. 2024.
\newblock A Survey on Large Language Models for Code Generation.
\newblock arXiv:2406.00515.

\bibitem[{Jimenez et~al.(2024)Jimenez, Yang, Wettig, Yao, Pei, Press, and Narasimhan}]{jimenez2024swebenchlanguagemodelsresolve}
Jimenez, C.~E.; Yang, J.; Wettig, A.; Yao, S.; Pei, K.; Press, O.; and Narasimhan, K. 2024.
\newblock SWE-bench: Can Language Models Resolve Real-World GitHub Issues?
\newblock arXiv:2310.06770.

\bibitem[{Jin et~al.(2025)Jin, Zeng, Yue, Yoon, Arik, Wang, Zamani, and Han}]{jin2025searchr1trainingllmsreason}
Jin, B.; Zeng, H.; Yue, Z.; Yoon, J.; Arik, S.; Wang, D.; Zamani, H.; and Han, J. 2025.
\newblock Search-R1: Training LLMs to Reason and Leverage Search Engines with Reinforcement Learning.
\newblock arXiv:2503.09516.

\bibitem[{Kaelbling, Littman, and Moore(1996)}]{kaelbling1996reinforcementlearningsurvey}
Kaelbling, L.~P.; Littman, M.~L.; and Moore, A.~W. 1996.
\newblock Reinforcement Learning: A Survey.
\newblock arXiv:cs/9605103.

\bibitem[{Lai et~al.(2022)Lai, Li, Wang, Zhang, Zhong, Zettlemoyer, tau Yih, Fried, Wang, and Yu}]{lai2022ds1000naturalreliablebenchmark}
Lai, Y.; Li, C.; Wang, Y.; Zhang, T.; Zhong, R.; Zettlemoyer, L.; tau Yih, S.~W.; Fried, D.; Wang, S.; and Yu, T. 2022.
\newblock DS-1000: A Natural and Reliable Benchmark for Data Science Code Generation.
\newblock arXiv:2211.11501.

\bibitem[{Lewis et~al.(2021)Lewis, Perez, Piktus, Petroni et~al.}]{lewis2021retrievalaugmentedgenerationknowledgeintensivenlp}
Lewis, P.; Perez, E.; Piktus, A.; Petroni, F.; et~al. 2021.
\newblock Retrieval-Augmented Generation for Knowledge-Intensive NLP Tasks.
\newblock arXiv:2005.11401.

\bibitem[{Li et~al.(2025)Li, Zhao, Zhao, Zeng, Wu et~al.}]{li2025tooldecathlonbenchmarkinglanguage}
Li, J.; Zhao, W.; Zhao, J.; Zeng, W.; Wu, H.; et~al. 2025.
\newblock The Tool Decathlon: Benchmarking Language Agents for Diverse, Realistic, and Long-Horizon Task Execution.
\newblock arXiv:2510.25726.

\bibitem[{Li, Zou, and Liu(2025)}]{li2025torlscalingtoolintegratedrl}
Li, X.; Zou, H.; and Liu, P. 2025.
\newblock ToRL: Scaling Tool-Integrated RL.
\newblock arXiv:2503.23383.

\bibitem[{Liu et~al.(2024{\natexlab{a}})Liu, Zheng, Qiao, Duan, Fei, Zhou, Zhang, Zhang, Lin, and Chen}]{liu2024mathbenchevaluatingtheoryapplication}
Liu, H.; Zheng, Z.; Qiao, Y.; Duan, H.; Fei, Z.; Zhou, F.; Zhang, W.; Zhang, S.; Lin, D.; and Chen, K. 2024{\natexlab{a}}.
\newblock MathBench: Evaluating the Theory and Application Proficiency of LLMs with a Hierarchical Mathematics Benchmark.
\newblock arXiv:2405.12209.

\bibitem[{Liu et~al.(2025{\natexlab{a}})Liu, Fan, Jiang et~al.}]{liu2025synlogicsynthesizingverifiablereasoning}
Liu, J.; Fan, Y.; Jiang, Z.; et~al. 2025{\natexlab{a}}.
\newblock SynLogic: Synthesizing Verifiable Reasoning Data at Scale for Learning Logical Reasoning and Beyond.
\newblock arXiv:2505.19641.

\bibitem[{Liu et~al.(2025{\natexlab{b}})Liu, Li, Zhang, Li, Chen, Ji et~al.}]{liu2025webexplorerexploreevolvetraining}
Liu, J.; Li, Y.; Zhang, C.; Li, J.; Chen, A.; Ji, K.; et~al. 2025{\natexlab{b}}.
\newblock WebExplorer: Explore and Evolve for Training Long-Horizon Web Agents.
\newblock arXiv:2509.06501.

\bibitem[{Liu et~al.(2023)Liu, Xia, Wang, and ZHANG}]{liu2023is}
Liu, J.; Xia, C.~S.; Wang, Y.; and ZHANG, L. 2023.
\newblock Is Your Code Generated by Chat{GPT} Really Correct? Rigorous Evaluation of Large Language Models for Code Generation.
\newblock In \emph{Thirty-seventh Conference on Neural Information Processing Systems}.

\bibitem[{Liu and Zhang(2025)}]{code-r1}
Liu, J.; and Zhang, L. 2025.
\newblock Code-R1: Reproducing R1 for Code with Reliable Rewards.

\bibitem[{Liu et~al.(2024{\natexlab{b}})Liu, Wang, Yin, Molchanov, Wang, Cheng, and Chen}]{liu2024doraweightdecomposedlowrankadaptation}
Liu, S.-Y.; Wang, C.-Y.; Yin, H.; Molchanov, P.; Wang, Y.-C.~F.; Cheng, K.-T.; and Chen, M.-H. 2024{\natexlab{b}}.
\newblock DoRA: Weight-Decomposed Low-Rank Adaptation.
\newblock arXiv:2402.09353.

\bibitem[{Liu et~al.(2025{\natexlab{c}})Liu, Pandit, Ye, Choi, and Durrett}]{liu2025codeupdatearenabenchmarkingknowledgeediting}
Liu, Z.~L.; Pandit, S.; Ye, X.; Choi, E.; and Durrett, G. 2025{\natexlab{c}}.
\newblock CodeUpdateArena: Benchmarking Knowledge Editing on API Updates.
\newblock arXiv:2407.06249.

\bibitem[{Longpre et~al.(2022)Longpre, Perisetla, Chen, Ramesh, DuBois, and Singh}]{longpre2022entitybasedknowledgeconflictsquestion}
Longpre, S.; Perisetla, K.; Chen, A.; Ramesh, N.; DuBois, C.; and Singh, S. 2022.
\newblock Entity-Based Knowledge Conflicts in Question Answering.
\newblock arXiv:2109.05052.

\bibitem[{Lu et~al.(2024)Lu, Lu, Lange, Foerster, Clune, and Ha}]{lu2024aiscientist}
Lu, C.; Lu, C.; Lange, R.~T.; Foerster, J.; Clune, J.; and Ha, D. 2024.
\newblock The {AI} {S}cientist: Towards Fully Automated Open-Ended Scientific Discovery.
\newblock \emph{arXiv preprint arXiv:2408.06292}.

\bibitem[{Luo et~al.(2025)Luo, Yang, Meng, Li, Zhou, and Zhang}]{luo2025empiricalstudycatastrophicforgetting}
Luo, Y.; Yang, Z.; Meng, F.; Li, Y.; Zhou, J.; and Zhang, Y. 2025.
\newblock An Empirical Study of Catastrophic Forgetting in Large Language Models During Continual Fine-tuning.
\newblock arXiv:2308.08747.

\bibitem[{Ma et~al.(2025)Ma, Zhuang, Xu, Jiang, Chen, and Guo}]{ma2025sqlr1trainingnaturallanguage}
Ma, P.; Zhuang, X.; Xu, C.; Jiang, X.; Chen, R.; and Guo, J. 2025.
\newblock SQL-R1: Training Natural Language to SQL Reasoning Model By Reinforcement Learning.
\newblock arXiv:2504.08600.

\bibitem[{Nijkamp et~al.(2023)Nijkamp, Pang, Hayashi, Tu, Wang, Zhou, Savarese, and Xiong}]{nijkamp2023codegenopenlargelanguage}
Nijkamp, E.; Pang, B.; Hayashi, H.; Tu, L.; Wang, H.; Zhou, Y.; Savarese, S.; and Xiong, C. 2023.
\newblock CodeGen: An Open Large Language Model for Code with Multi-Turn Program Synthesis.
\newblock arXiv:2203.13474.

\bibitem[{OpenAI(2024)}]{gpt-o1}
OpenAI. 2024.
\newblock Introducing OpenAI o1.
\newblock \url{https://openai.com/o1/}.

\bibitem[{OpenAI et~al.(2024)OpenAI, Achiam, Adler, Agarwal, Ahmad et~al.}]{openai2024gpt4technicalreport}
OpenAI; Achiam, J.; Adler, S.; Agarwal, S.; Ahmad, L.; et~al. 2024.
\newblock GPT-4 Technical Report.
\newblock arXiv:2303.08774.

\bibitem[{Ouyang et~al.(2022)Ouyang, Wu, Jiang, Almeida, Wainwright, Mishkin, Zhang, Agarwal, Slama, Ray, Schulman, Hilton, Kelton, Miller, Simens, Askell, Welinder, Christiano, Leike, and Lowe}]{ouyang2022traininglanguagemodelsfollow}
Ouyang, L.; Wu, J.; Jiang, X.; Almeida, D.; Wainwright, C.~L.; Mishkin, P.; Zhang, C.; Agarwal, S.; Slama, K.; Ray, A.; Schulman, J.; Hilton, J.; Kelton, F.; Miller, L.; Simens, M.; Askell, A.; Welinder, P.; Christiano, P.; Leike, J.; and Lowe, R. 2022.
\newblock Training language models to follow instructions with human feedback.
\newblock arXiv:2203.02155.

\bibitem[{Qian et~al.(2025)Qian, Acikgoz, He, Wang, Chen, Hakkani-Tür, Tur, and Ji}]{qian2025toolrlrewardtoollearning}
Qian, C.; Acikgoz, E.~C.; He, Q.; Wang, H.; Chen, X.; Hakkani-Tür, D.; Tur, G.; and Ji, H. 2025.
\newblock ToolRL: Reward is All Tool Learning Needs.
\newblock arXiv:2504.13958.

\bibitem[{Rafailov et~al.(2024)Rafailov, Sharma, Mitchell, Ermon, Manning, and Finn}]{rafailov2024directpreferenceoptimizationlanguage}
Rafailov, R.; Sharma, A.; Mitchell, E.; Ermon, S.; Manning, C.~D.; and Finn, C. 2024.
\newblock Direct Preference Optimization: Your Language Model is Secretly a Reward Model.
\newblock arXiv:2305.18290.

\bibitem[{Rozière et~al.(2024)Rozière, Gehring, Gloeckle, Sootla et~al.}]{rozière2024codellamaopenfoundation}
Rozière, B.; Gehring, J.; Gloeckle, F.; Sootla, S.; et~al. 2024.
\newblock Code Llama: Open Foundation Models for Code.
\newblock arXiv:2308.12950.

\bibitem[{Schulman et~al.(2017)Schulman, Wolski, Dhariwal, Radford, and Klimov}]{schulman2017proximalpolicyoptimizationalgorithms}
Schulman, J.; Wolski, F.; Dhariwal, P.; Radford, A.; and Klimov, O. 2017.
\newblock Proximal Policy Optimization Algorithms.
\newblock arXiv:1707.06347.

\bibitem[{Shao et~al.(2024)Shao, Wang, Zhu, Xu, Song, Bi, Zhang, Zhang, Li, Wu, and Guo}]{shao2024deepseekmathpushinglimitsmathematical}
Shao, Z.; Wang, P.; Zhu, Q.; Xu, R.; Song, J.; Bi, X.; Zhang, H.; Zhang, M.; Li, Y.~K.; Wu, Y.; and Guo, D. 2024.
\newblock DeepSeekMath: Pushing the Limits of Mathematical Reasoning in Open Language Models.
\newblock arXiv:2402.03300.

\bibitem[{Starace et~al.(2025)Starace, Jaffe, Sherburn, Aung, Chan, Maksin, Dias, Mays, Kinsella, Thompson, Heidecke, Glaese, and Patwardhan}]{starace2025paperbenchevaluatingaisability}
Starace, G.; Jaffe, O.; Sherburn, D.; Aung, J.; Chan, J.~S.; Maksin, L.; Dias, R.; Mays, E.; Kinsella, B.; Thompson, W.; Heidecke, J.; Glaese, A.; and Patwardhan, T. 2025.
\newblock PaperBench: Evaluating AI's Ability to Replicate AI Research.
\newblock arXiv:2504.01848.

\bibitem[{Team et~al.(2024)Team, Zhao, Hui, Howland, Nguyen, Zuo et~al.}]{codegemmateam2024codegemmaopencodemodels}
Team, C.; Zhao, H.; Hui, J.; Howland, J.; Nguyen, N.; Zuo, S.; et~al. 2024.
\newblock CodeGemma: Open Code Models Based on Gemma.
\newblock arXiv:2406.11409.

\bibitem[{Team et~al.(2025{\natexlab{a}})Team, Anil, Borgeaud, Alayrac et~al.}]{geminiteam2025geminifamilyhighlycapable}
Team, G.; Anil, R.; Borgeaud, S.; Alayrac, J.-B.; et~al. 2025{\natexlab{a}}.
\newblock Gemini: A Family of Highly Capable Multimodal Models.
\newblock arXiv:2312.11805.

\bibitem[{Team et~al.(2025{\natexlab{b}})Team, Du, Gao, Xing, Jiang, Chen, Li, Xiao et~al.}]{kimiteam2025kimik15scalingreinforcement}
Team, K.; Du, A.; Gao, B.; Xing, B.; Jiang, C.; Chen, C.; Li, C.; Xiao, C.; et~al. 2025{\natexlab{b}}.
\newblock Kimi k1.5: Scaling Reinforcement Learning with LLMs.
\newblock arXiv:2501.12599.

\bibitem[{Team et~al.(2025{\natexlab{c}})Team, Li, Zhang, Zhang, Huang et~al.}]{tongyideepresearchteam2025tongyideepresearchtechnicalreport}
Team, T.~D.; Li, B.; Zhang, B.; Zhang, D.; Huang, F.; et~al. 2025{\natexlab{c}}.
\newblock Tongyi DeepResearch Technical Report.
\newblock arXiv:2510.24701.

\bibitem[{Wang et~al.(2025{\natexlab{a}})Wang, Huang, Zhang, Feng, Zhang, Liu, and Peng}]{wang2025llmsmeetlibraryevolution}
Wang, C.; Huang, K.; Zhang, J.; Feng, Y.; Zhang, L.; Liu, Y.; and Peng, X. 2025{\natexlab{a}}.
\newblock LLMs Meet Library Evolution: Evaluating Deprecated API Usage in LLM-based Code Completion.
\newblock arXiv:2406.09834.

\bibitem[{Wang et~al.(2025{\natexlab{b}})Wang, Qian, Zhong, Chen, Qiu, Huang, Jin, Wang, Wong, and Ji}]{wang2025otcoptimaltoolcalls}
Wang, H.; Qian, C.; Zhong, W.; Chen, X.; Qiu, J.; Huang, S.; Jin, B.; Wang, M.; Wong, K.-F.; and Ji, H. 2025{\natexlab{b}}.
\newblock OTC: Optimal Tool Calls via Reinforcement Learning.
\newblock arXiv:2504.14870.

\bibitem[{Wang et~al.(2025{\natexlab{c}})Wang, Tian, Wang, Zhang, Huang, Wu, and Jiang}]{wang2025simplearpushingfrontierautoregressive}
Wang, J.; Tian, Z.; Wang, X.; Zhang, X.; Huang, W.; Wu, Z.; and Jiang, Y.-G. 2025{\natexlab{c}}.
\newblock SimpleAR: Pushing the Frontier of Autoregressive Visual Generation through Pretraining, SFT, and RL.
\newblock arXiv:2504.11455.

\bibitem[{Wang et~al.(2025{\natexlab{d}})Wang, Danek, Yang, Chen, and Sun}]{wang2025largelanguagemodelsreplace}
Wang, Z.; Danek, B.; Yang, Z.; Chen, Z.; and Sun, J. 2025{\natexlab{d}}.
\newblock Can Large Language Models Replace Data Scientists in Biomedical Research?
\newblock arXiv:2410.21591.

\bibitem[{Wu et~al.(2025)Wu, Wang, Zhao, and He}]{wu2025miragemethodmodeltaskalignment}
Wu, H.; Wang, C.; Zhao, W.; and He, J. 2025.
\newblock Mirage or Method? How Model-Task Alignment Induces Divergent RL Conclusions.
\newblock arXiv:2508.21188.

\bibitem[{Wu et~al.(2024)Wu, Wu, Wang, Xu, Ma, Jiang, Yang, Xing, Li, and Haffari}]{wu2024versicodeversioncontrollablecodegeneration}
Wu, T.; Wu, W.; Wang, X.; Xu, K.; Ma, S.; Jiang, B.; Yang, P.; Xing, Z.; Li, Y.-F.; and Haffari, G. 2024.
\newblock VersiCode: Towards Version-controllable Code Generation.
\newblock arXiv:2406.07411.

\bibitem[{Xie et~al.(2025)Xie, Gao, Ren, Luo, Hong, Dai, Zhou, Qiu, Wu, and Luo}]{xie2025logicrlunleashingllmreasoning}
Xie, T.; Gao, Z.; Ren, Q.; Luo, H.; Hong, Y.; Dai, B.; Zhou, J.; Qiu, K.; Wu, Z.; and Luo, C. 2025.
\newblock Logic-RL: Unleashing LLM Reasoning with Rule-Based Reinforcement Learning.
\newblock arXiv:2502.14768.

\bibitem[{Xu et~al.(2024)Xu, Qi, Guo, Wang, Wang, Zhang, and Xu}]{xu2024knowledgeconflictsllmssurvey}
Xu, R.; Qi, Z.; Guo, Z.; Wang, C.; Wang, H.; Zhang, Y.; and Xu, W. 2024.
\newblock Knowledge Conflicts for LLMs: A Survey.
\newblock arXiv:2403.08319.

\bibitem[{Yang, Li et~al.(2025)}]{qwen3}
Yang, A.; Li, A.; et~al. 2025.
\newblock Qwen3 Technical Report.
\newblock \emph{arXiv preprint arXiv:2505.09388}.

\bibitem[{Yao et~al.(2023)Yao, Wang, Tian, Cheng, Li, Deng, Chen, and Zhang}]{yao2023editing}
Yao, Y.; Wang, P.; Tian, B.; Cheng, S.; Li, Z.; Deng, S.; Chen, H.; and Zhang, N. 2023.
\newblock Editing large language models: Problems, methods, and opportunities.
\newblock \emph{arXiv preprint arXiv:2305.13172}.

\bibitem[{Yenduri et~al.(2023)Yenduri, M, G, Y, Srivastava et~al.}]{yenduri2023generativepretrainedtransformercomprehensive}
Yenduri, G.; M, R.; G, C.~S.; Y, S.; Srivastava, G.; et~al. 2023.
\newblock Generative Pre-trained Transformer: A Comprehensive Review on Enabling Technologies, Potential Applications, Emerging Challenges, and Future Directions.
\newblock arXiv:2305.10435.

\bibitem[{Yu et~al.(2025)Yu, Zhang, Zhu, Yuan, Zuo, Yue, Fan, Liu, Liu, Liu, Lin, Lin, Ma, Sheng, Tong, Zhang, Zhang, Zhang, Zhu, Zhu, Chen, Chen, Wang, Yu, Dai, Song, Wei, Zhou, Liu, Ma, Zhang, Yan, Qiao, Wu, and Wang}]{yu2025dapoopensourcellmreinforcement}
Yu, Q.; Zhang, Z.; Zhu, R.; Yuan, Y.; Zuo, X.; Yue, Y.; Fan, T.; Liu, G.; Liu, L.; Liu, X.; Lin, H.; Lin, Z.; Ma, B.; Sheng, G.; Tong, Y.; Zhang, C.; Zhang, M.; Zhang, W.; Zhu, H.; Zhu, J.; Chen, J.; Chen, J.; Wang, C.; Yu, H.; Dai, W.; Song, Y.; Wei, X.; Zhou, H.; Liu, J.; Ma, W.-Y.; Zhang, Y.-Q.; Yan, L.; Qiao, M.; Wu, Y.; and Wang, M. 2025.
\newblock DAPO: An Open-Source LLM Reinforcement Learning System at Scale.
\newblock arXiv:2503.14476.

\bibitem[{Yue et~al.(2025)Yue, Chen, Lu, Zhao, Wang, Yue, Song, and Huang}]{yue2025doesreinforcementlearningreally}
Yue, Y.; Chen, Z.; Lu, R.; Zhao, A.; Wang, Z.; Yue, Y.; Song, S.; and Huang, G. 2025.
\newblock Does Reinforcement Learning Really Incentivize Reasoning Capacity in LLMs Beyond the Base Model?
\newblock arXiv:2504.13837.

\bibitem[{Zan et~al.(2023)Zan, Chen, Zhang, Lu, Wu, Guan, Wang, and Lou}]{zan2023largelanguagemodelsmeet}
Zan, D.; Chen, B.; Zhang, F.; Lu, D.; Wu, B.; Guan, B.; Wang, Y.; and Lou, J.-G. 2023.
\newblock Large Language Models Meet NL2Code: A Survey.
\newblock arXiv:2212.09420.

\bibitem[{Zeng et~al.(2025)Zeng, Huang, Liu, Liu, He, Ma, and He}]{zeng2025simplerlzooinvestigatingtamingzero}
Zeng, W.; Huang, Y.; Liu, Q.; Liu, W.; He, K.; Ma, Z.; and He, J. 2025.
\newblock SimpleRL-Zoo: Investigating and Taming Zero Reinforcement Learning for Open Base Models in the Wild.
\newblock arXiv:2503.18892.

\bibitem[{Zhang et~al.(2023)Zhang, Fang, Xie, Zhang, Yang, Sun, Yu, and Chen}]{zhang2023survey}
Zhang, Q.; Fang, C.; Xie, Y.; Zhang, Y.; Yang, Y.; Sun, W.; Yu, S.; and Chen, Z. 2023.
\newblock A Survey on Large Language Models for Software Engineering.
\newblock \emph{arXiv preprint arXiv:2312.15223}.

\bibitem[{Zhang et~al.(2024)Zhang, Chen, Liu, Liao, Gong, Yu, Li, and Wang}]{zhang2024unifying}
Zhang, Z.; Chen, C.; Liu, B.; Liao, C.; Gong, Z.; Yu, H.; Li, J.; and Wang, R. 2024.
\newblock Unifying the Perspectives of {NLP} and Software Engineering: A Survey on Language Models for Code.
\newblock \emph{Transactions on Machine Learning Research}.

\bibitem[{Zheng et~al.(2024)Zheng, Xia, Zou, Dong, Wang, Xue, Wang, Shen, Wang, Li, Su, Yang, and Tang}]{zheng2024codegeexpretrainedmodelcode}
Zheng, Q.; Xia, X.; Zou, X.; Dong, Y.; Wang, S.; Xue, Y.; Wang, Z.; Shen, L.; Wang, A.; Li, Y.; Su, T.; Yang, Z.; and Tang, J. 2024.
\newblock CodeGeeX: A Pre-Trained Model for Code Generation with Multilingual Benchmarking on HumanEval-X.
\newblock arXiv:2303.17568.

\bibitem[{Zhuo et~al.(2025)Zhuo, Vu, Chim, Hu, Yu, Widyasari et~al.}]{zhuo2025bigcodebenchbenchmarkingcodegeneration}
Zhuo, T.~Y.; Vu, M.~C.; Chim, J.; Hu, H.; Yu, W.; Widyasari, R.; et~al. 2025.
\newblock BigCodeBench: Benchmarking Code Generation with Diverse Function Calls and Complex Instructions.
\newblock arXiv:2406.15877.

\end{thebibliography}

\newpage



\appendix

\section{Context of Reinforcement Learning}
\label{app:rl}

Reinforcement Learning (RL) training remains an important post-training component for LLMs.
Reinforcement Learning from Human Feedback (RLHF) is a key contributor to the upgrade of GPT to ChatGPT~\cite{ouyang2022traininglanguagemodelsfollow}.
By training a reward model that aligns with human preferences to provide reward signals for RL training, it makes LLMs more interactive.
Subsequent algorithms like Direct Preference Optimization (DPO)~\cite{rafailov2024directpreferenceoptimizationlanguage} do not explicitly use RL algorithms, but the underlying idea is consistent with RLHF.
With the emergence of slow-thinking reasoning models~\cite{gpt-o1,deepseekai2025deepseekr1incentivizingreasoningcapability,kimiteam2025kimik15scalingreinforcement}, the community has recognized the importance of rule-based RL.
Compared to SFT, rule-based Reinforcement Fine-Tuning (rule-based RFT) demonstrates stronger generalization capabilities~\cite{chu2025sftmemorizesrlgeneralizes}.
Particularly, the "Aha moments" observed in mathematical and logical tasks are quite fascinating~\cite{deepseekai2025deepseekr1incentivizingreasoningcapability,zeng2025simplerlzooinvestigatingtamingzero,wu2025miragemethodmodeltaskalignment}.
Recently, the community has found that RFT is also effective in other domain tasks, such as video understanding~\cite{feng2025videor1reinforcingvideoreasoning}, image generation~\cite{wang2025simplearpushingfrontierautoregressive}, code generation~\cite{code-r1,ma2025sqlr1trainingnaturallanguage}, tool using~\cite{li2025torlscalingtoolintegratedrl,qian2025toolrlrewardtoollearning,wang2025otcoptimaltoolcalls}, machine translation~\cite{feng2025mtr1zeroadvancingllmbasedmachine}, and others~\cite{jin2025searchr1trainingllmsreason}.

At the same time, the underlying reinforcement learning algorithms used for model optimization are also being continuously updated.
PPO (Proximal Policy Optimization)~\cite{schulman2017proximalpolicyoptimizationalgorithms} algorithm is one of the most popular RL algorithms in recent years.
An additional model of criticism needs to be trained to predict the advantages of the actions taken.
GRPO (Group Reward Policy Optimization) algorithm,  introduced by DeepSeek-MATH~\cite{shao2024deepseekmathpushinglimitsmathematical}, leverages group-wise advantages as rewards, eliminating the need for the critic model in PPO.
Specifically, the training objective of GRPO is:
\begin{equation}
\begin{split}
&{\arg\max}_\theta J_{GRPO}(\theta)= \\
&\mathbb{E}_{q\sim P(Q), \{o_i\}_{i=1}^G\sim \pi_{\theta_k}(O|q)}
\frac{1}{G}\sum_{i=1}^G\frac{1}{|o_i|}\sum_{t=1}^{|o_i|} \\
&\{\min[\frac{\pi_{\theta}(o_{i,t}|q,o_{i,<t})}{\pi_{\theta_k}(o_{i,t}|q,o_{i,<t})}\hat{A}_{i,t},\text{clip}(\frac{\pi_{\theta}(o_{i,t}|q,o_{i,<t})}{\pi_{\theta_k}(o_{i,t}|q,o_{i,<t})}, \\
&1-\epsilon,1+\epsilon)\hat{A}_{i,t}]
-\beta D_{KL}[\pi_\theta|\pi_{ref}]\}
\end{split}
\end{equation}
\noindent where $\pi_\theta$ is the policy model (LLM to be optimized), $\pi_{ref}$ is the reference model, $\pi_{\theta_k}$ is the old policy model, $Q$ is the question dataset.
GRPO algorithm simultaneously generates multiple responses from the policy model $\{o_i\}_{i=1}^G$ and calculates rewards $\{r_i\}_{i=1}^G$, thereby obtaining the inter-group advantage as the advantage function:
\begin{equation}
    \hat{A}_{i,t}=\hat{A}_i=\frac{r_i-\text{avg}(\{r_i\}_{i=1}^N\})}{\text{std}(\{r_i\}_{i=1}^N\})}
    \label{eq:naive-GRPO}
\end{equation}
DAPO (Dynamic Advantage Policy Optimization) algorithm~\cite{yu2025dapoopensourcellmreinforcement} analyzes the shortcomings in GRPO and makes improvements, filling the gaps in the details of the GRPO algorithm as mentioned in the DeepSeek-R1 technical report~\cite{deepseekai2025deepseekr1incentivizingreasoningcapability}.
DAPO modifies the training objective to:
\begin{equation}
\begin{split}
    &{\arg\max}_\theta{J}_{DAPO}(\theta) = \\
    &\mathbb{E}_{(q,a)\sim \mathcal{D}, \{o_i\}_{i=1}^G\sim \pi_k(\cdot|q)}[\frac{1}{\sum_{i=1}^G|o_i|}\sum_{i=1}^G\sum_{t=1}^{|o_i|} min( \\
    &r_{i,t}(\theta) \hat{A}_{i,t},\text{clip}(r_{i,t}(\theta),1-\epsilon_{low},1+\epsilon_{high})\hat{A}_{i,t})]\\
    &s.t. \,\,0<|\{o_i|is\_equivalent(a,o_i)\}|<G
\end{split}
\label{eq:dapo-obj}
\end{equation}
where $r_{i,t}(\theta)=\frac{\pi_\theta(o_{i,t}|q,o_{i,<t})}{\pi_k(o_{i,t}|q,o_{i,<t})}$.
It includes three modifications: 1) the KL divergence is removed, 2) the upper bound of the clip is relaxed $\epsilon_{high} > \epsilon_{low})$, 3) using token-level policy gradient Loss and 4) the inter-group advantage being zero is avoided by dynamic sampling.
DAPO also takes into account that reward noise is introduced when the output length exceeds the set maximum length.
Therefore, the algorithm also introduces an additional length penalty:
\begin{equation}
\begin{split}
R_{len}(x)=\left\{\begin{matrix}
 &0.0,|x| \le L_{max}-L_{cache}\\
&-1.0,|x|>L_{max} \\
 &\frac{(L_{max}-L_{cache})-|y|}{L_{cache}}, otherwise
\end{matrix}\right.
\end{split}
\label{eq:dapo-pen}
\end{equation}

\section{Collected Dataset}
\label{app:stat}

\subsection{Human Expert Review}

We recruit five students with a background in computer software to review the generated code.
They focus on checking whether the generated content includes the updated API and on verifying the code's correctness.
When errors occur, we provide GPT-4 with human feedback about the errors and request it to regenerate the content again.

\subsection{Statistics}
Table~\ref{tab:statistics} presents the dataset statistics we collected.
All data are collected from mainstream data science libraries.

\begin{table*}[!ht]
    \centering
    \begin{tabular}{c|c|c}
    \toprule
        \textbf{\,\,\,\,\,\,\,\,Library\,\,\,\,\,\,\,\,} & \textbf{\,\,\,\,\,\,\,\,Versions\,\,\,\,\,\,\,\,} & \textbf{Number of Data Entries} \\
    \midrule
        NumPy & 1.3.0 - 2.2.0 & 199 \\
        Pandas & 0.24.0 - 2.2.3 & 243 \\
        PyTorch & 1.0.0 - 2.6.0 & 273 \\
        Matplotlib & 1.0 - 3.10 & 175 \\
        Scikit-Learn & 0.13 - 1.6 & 192 \\
        Scipy & 0.10.0 - 1.15.0 & 203 \\
        Scrapy & 1.0.0 - 2.11.0 & 274 \\
        Seaborn & 0.8.0 - 0.13.2 & 142 \\
        Jax & 0.3.5 - 0.6.0 & 85 \\
    \midrule
        Total & & 1786 \\
    \bottomrule
    \end{tabular}
    \caption{Statistics of our collected dataset.}
    \label{tab:statistics}
\end{table*}

The diversity of our dataset is not only reflected in the variety of libraries, but also in the diversity of API update types.
In addition to common API updates such as name changes and added parameters, our dataset also includes a wider range of update types. 
Several examples are shown below:

\begin{lstlisting}[language={python}]
# Example 1
# _add_newdoc_ufunc is now deprecated. 
# ufunc.doc = newdoc should be used.
# Old Code
import numpy as np

f = np.frompyfunc(lambda x: x**2,1,1)
np._add_newdoc_ufunc(f, "A custom \
    ufunc that squares the input.")

# New Code
import numpy as np

f = np.frompyfunc(lambda x: x**2,1,1)
f.__doc__ = "A custom ufunc that \
    squares the input."


# Example 2
# Arrays of 2-dimensional vectors for 
# np.cross have been deprecated. 
# Use arrays of 3-dimensional vectors.
# Old Code
import numpy as np

a = np.array([1, 2])
b = np.array([3, 4])
result = np.cross(a, b)

# New Code
import numpy as np

a = np.array([1, 2, 0])
b = np.array([3, 4, 0])
result = np.cross(a, b)
\end{lstlisting}

\section{Prompt with Template}
\label{app:prompt}

The prompt template we use during training is as follows:

\begin{prompt}[title={\thetcbcounter {} Training Task (Code Migration) Prompt Template}]

\promptsubsection{System}\\
You are a helpful coding assistant. 
Your task is to transform the old version of the code into the new version specified, based on the update information. You first thinks about the reasoning process in the mind and then provides the solution.\\
\\
\promptsubsection{User}\\
Dependency $d_i$ performed an API update in version $v_i$, and the update content includes:

<doc>

{\centering update info $u_i$ \par}

</doc> \\
The old version of the code is:

```python

{\centering old code $c_i^{(old)}$ \par}

```

Show your work in <think> </think> tags. And return the final code in <answer> </answer>, the code within <answer></answer> should be enclosed in ```python ``` tags.

\promptsubsection{Assistant}\\
Let me solve this step by step. \\
<think>

\end{prompt}

\noindent And the prompt template we use during testing is as follows:

\begin{prompt}[title={\thetcbcounter {} Testing Task (CodeUpdateArena) Prompt Template}]

\promptsubsection{System}\\
You are a helpful code assistant. 
You first think about the reasoning process in the mind and then provide a Python solution to a problem in a real-world scenario.

\promptsubsection{User}\\
Update Note:

There's an recent update to a function $u_i$['update\_api\_path'] -- $u_i$['update\_description'].

The function now has a new function signature -- $u_i$['update\_signature'].

Here is a detailed documentation about the update:

<doc>

{\centering $u_i$['update\_docstring'] \par}

</doc>

Scenario: $q_i$['Scenario']

Problem: $q_i$['problem']

Solution Signature: $q_i$['signature']

Show your work in <think> </think> tags.
And return the final code in <answer> </answer>, the code within <answer></answer> should be enclosed in ```python ``` tags.

\promptsubsection{Assistant}\\
Let me solve this step by step. \\
<think>

\end{prompt}

\section{Baseline Selection}
\label{app:baseline}

As far as we know, there is no other community-recognized baseline for training on code migration tasks and testing on CodeUpdateArena~\cite{liu2025codeupdatearenabenchmarkingknowledgeediting}.
Many rule-based RFT works do not set other baselines, usually focusing more on whether the training can bring improvements or surpass a reasoning model with more parameters~\cite{yu2025dapoopensourcellmreinforcement,liu2025synlogicsynthesizingverifiablereasoning,feng2025mtr1zeroadvancingllmbasedmachine}.
We also adopt this setting and choose two different 32B models to serve as the baseline:

\begin{itemize}
    \item Qwen2.5-Coder-32B-Instruct: it has stronger coding capabilities.
    \item DeepSeek-R1-Distill-Qwen-32B: it has stronger reasoning capabilities.
\end{itemize}

\section{Test Dataset}
\label{app:test-data}

When using the CodeUpdateArena dataset, we find that some of the test cases have issues.
In our experiments, we make modifications to these unreasonable aspects.
Here are the three examples that we have identified:

\textbf{Example 1: In some test cases, two dictionaries are directly compared using == to check if they are the same}.

\begin{lstlisting}[language={python}]
assert result == expected_result
\end{lstlisting}

\noindent We replace such test cases with correct Python code to compare two dictionaries.

\textbf{Example 2: The role of a certain parameter in the solution signature is not clear}.
The solution function signature to be implemented is:
\begin{lstlisting}[language={python}]
def convert_gradian_to_degree(
    angle_list: List[float], 
    flag_list: List[bool]
):
    pass
\end{lstlisting}

\noindent However, the meaning of the flag parameter is not clearly specified in the update prompt.
We supplement the meaning of the flag in the prompt based on the logic of the test cases.

\textbf{Example 3: The value of the solution parameter is in conflict with the test case.}
In the update prompt, the newly introduced angle unit values are specified as 'radians' and 'degrees', but the test cases use the singular form.

\begin{lstlisting}[language={python}]
def test_angles_in_degrees():
    angle_sequence = [0, 60, 180, 300]
    unit = 'degree'  # Conflict
    result = calculate_sine_of_angles(
        angle_sequence, unit
    )
    import math
    
    expected_results = []
    
    for angle in angle_sequence:
        rad = math.radians(angle)
        expected_results.append(
            math.sin(rad)
        )
    
    for i in range(len(result)):
        assert math.isclose(
            result[I], 
            expected_results[I], 
            rel_tol=1e-9
        )
\end{lstlisting}
We unify the singular and plural forms of the units.

After correcting these errors, the Pass@1 rate of DS-Coder-1.5 increases from 46.8 (value in CodeUpdateArena~\cite{liu2025codeupdatearenabenchmarkingknowledgeediting} paper) to 59.1.

\section{Case Study}

\subsection{Error case 1}
\label{app:err-case-1}

\begin{prompt}[title={\thetcbcounter {} Error Case}]

\promptsubsection{System}\\
You are a helpful code assistant. 
You first think about the reasoning process in the mind and then provide a Python solution to a problem in a real-world scenario.

\promptsubsection{User}\\
Update Note:

There's an recent update to a function math.degrees -- The math.degrees function is updated to enable conversion from radians and another unit through an additional parameter.

The function now has a new function signature -- math.degrees(x, input\_unit='radian').

Here is a detailed documentation about the update:

<doc>

The function now includes an additional parameter 'input\_unit' which defaults to 'radian'. The 'input\_unit' can be either 'radian' or 'gradian'. When 'input\_unit' is 'radian', the function converts the specified radian into degrees. When 'input\_unit' is 'gradian', the function converts the specified gradian into degrees. If the input\_unit is neither 'radian' nor 'gradian', a ValueError is raised with the message 'input\_unit should be either radian or gradian'.

</doc>

Scenario: You are working on a scientific computing project, where angle calculations are a crucial part. However, not all your data sources provide angles in the same measurement units, some are providing angle values in radians and others in gradians. You are aiming to standardize all angles to degrees for improved interoperability.

Problem: Given an angle value and the unit it's provided in (either 'radian' or 'gradian'), write a function to convert it into degrees.

Solution signature: function convert\_angle\_to\_degrees(angle: float, unit: str) -> float:

Show your work in <think> </think> tags.
And return the final code in <answer> </answer>, the code within <answer></answer> should be enclosed in ```python ``` tags.

\promptsubsection{Assistant}\\
Let me solve this step by step. \\
<think>

1. First, I need to understand how to convert radians to degrees. The formula is degrees = radians * (180 / $\pi$).

2. Next, I need to understand how to convert gradians to degrees. The formula is \textcolor{red}{degrees = gradians * (90 / 200)}.

3. Now, I need to create a function that takes two parameters: the angle value and its unit ('radian' or 'gradian').

4. Inside the function, I'll check the unit parameter. If it's 'radian', I'll use the radians to degrees formula. If it's 'gradian', I'll use the gradians to degrees formula.

5. If the unit parameter is neither 'radian' nor 'gradian', I'll raise a ValueError with the appropriate message.

6. Finally, I'll return the converted angle value.

</think>

<answer>
\begin{lstlisting}[language={python}]
import math

def convert_angle_to_degrees(
    angle: float, unit: str
) -> float:
    if unit == 'radian':
        return angle * (180 / math.pi)
    elif unit == 'gradian':
        return angle * (90 / 200)
    else:
        raise ValueError('input_unit \
            should be either radian \
            or gradian')
\end{lstlisting}
</answer>
\end{prompt}

\subsection{Correct Case}
\label{app:correct-case}

\begin{prompt}[title={\thetcbcounter {} Correct Case}]

\promptsubsection{System}\\
You are a helpful code assistant. 
You first think about the reasoning process in the mind and then provide a Python solution to a problem in a real-world scenario.

\promptsubsection{User}\\
Update Note:

There's an recent update to a function math.degrees -- The math.degrees function is updated to enable conversion from radians and another unit through an additional parameter.

The function now has a new function signature -- math.degrees(x, input\_unit='radian').

Here is a detailed documentation about the update:

<doc>

The function now includes an additional parameter 'input\_unit' which defaults to 'radian'. The 'input\_unit' can be either 'radian' or 'gradian'. When 'input\_unit' is 'radian', the function converts the specified radian into degrees. When 'input\_unit' is 'gradian', the function converts the specified gradian into degrees. If the input\_unit is neither 'radian' nor 'gradian', a ValueError is raised with the message 'input\_unit should be either radian or gradian'.

</doc>

Scenario: You are working on a scientific computing project, where angle calculations are a crucial part. However, not all your data sources provide angles in the same measurement units, some are providing angle values in radians and others in gradians. You are aiming to standardize all angles to degrees for improved interoperability.

Problem: Given an angle value and the unit it's provided in (either 'radian' or 'gradian'), write a function to convert it into degrees.

Solution signature: function convert\_angle\_to\_degrees(angle: float, unit: str) -> float:

Show your work in <think> </think> tags.
And return the final code in <answer> </answer>, the code within <answer></answer> should be enclosed in ```python ``` tags.

\promptsubsection{Assistant}\\
Let me solve this step by step. \\
<think>

To solve this problem, we need to create a function that leverages the updated `math.degrees` function to convert angles from either radians or gradians to degrees. The function will take two parameters: the angle value and the unit of that angle ('radian' or 'gradian').

Here's the step-by-step thought process:

1. We will define a function named convert\_angle\_to\_degrees that takes two arguments: angle (a float representing the angle value) and unit (a string indicating whether the angle is in 'radian' or 'gradian').

2.Inside the function, we will call the updated math.degrees function, passing the angle and unit as arguments. This function will handle the conversion based on the unit provided.

3. Since the math.degrees function already raises a ValueError if the input\_unit is not 'radian' or 'gradian', we don't need to add any additional error handling in our function.

4. Finally, we will return the result of the math.degrees function call, which will be the angle converted to degrees.
We'll also import the math module at the beginning of our script since we're using its degrees function.

</think>

<answer>
\begin{lstlisting}[language={python}]
import math

def convert_angle_to_degrees(
    angle: float, unit: str
) -> float:
    return math.degrees(angle, unit)

\end{lstlisting}
</answer>

\end{prompt}

\subsection{Error case 2}
\label{app:error-case-2}

When simulating a pendulum, the model lacks an understanding of physical knowledge. 
In the generated code, it completely ignores the physical quantities that need to be used.

\begin{lstlisting}[language={python}]
def get_pendulum_position(
    angle_in_degrees, gravity, 
    length, time
):
    # Convert from degrees to radians
    angle_in_radians = math.radians(
        angle_in_degrees
    )
    
    # Calculate the sine of the angle
    sine_value = math.sin(
        angle_in_radians
    )
    
    # Calculate the pendulum position
    position = angle_in_degrees * \
        sine_value
    
    return position

\end{lstlisting}

\section{Results under Complex Scenarios}

\subsection{Multiple updates to a single API}
\label{app:mul-app}

\subsection{Multiple APIs Update}
\label{app:mul-api}
\begin{prompt}[title={\thetcbcounter {} Error Case}]

\promptsubsection{System}\\
You are a helpful code assistant. 
You first think about the reasoning process in the mind and then provide a Python solution to a problem in a real-world scenario.

\promptsubsection{User}\\
Update Note:

There's an recent update to a function math.degrees -- The math.degrees function is updated to enable conversion from radians and another unit through an additional parameter.

The function now has a new function signature -- math.degrees(x, input\_unit='radian').

Here is a detailed documentation about the update:

<doc>

The function now includes an additional parameter 'input\_unit' which defaults to 'radian'. The 'input\_unit' can be either 'radian' or 'gradian'. When 'input\_unit' is 'radian', the function converts the specified radian into degrees. When 'input\_unit' is 'gradian', the function converts the specified gradian into degrees. If the input\_unit is neither 'radian' nor 'gradian', a ValueError is raised with the message 'input\_unit should be either radian or gradian'.

</doc>

There's an other update to a function math.log -- Add a boolean parameter 'inverse' to math.log to control whether to perform logarithm or anti-logarithm calculation.

The function now has a new function signature -- math.log(x, base=math.e, inverse=False)

<doc>

An optional third parameter 'inverse' has been added to the 'log' function. If 'inverse' is set to True, the function will instead calculate the anti-logarithm (exponentiation) of 'x' to the base 'base'. The value of 'inverse' defaults to False, i.e., the function will perform a logarithm calculation as before if 'inverse' is not explicitly set to True.

Note: If 'inverse' is set to True and 'base' is not specified, the function will calculate the exponential of 'x'.

</doc>

Scenario: John has a angle 'x' (with a specific unit) and a base 'b'. He also has a boolean flag 'flag'. If 'flag' is true, he wants to perform an anti-logarithm operation with base 'b' on the degree of angle 'x'. If 'flag' is false, he wants to perform a logarithm operation with base 'b' on the degree of 'x'. He needs a Python function that can perform these operations for him based on the values of 'x', 'b' and 'flag'.

Problem: Given an angle value and the unit it's provided in (either 'radian' or 'gradian'), write a function to convert it into log or anti-log.

Solution signature: def convert\_angle\_to\_log\_or\_antilog(x, b, flag, input\_unit):

Show your work in <think> </think> tags.
And return the final code in <answer> </answer>, the code within <answer></answer> should be enclosed in ```python ``` tags.

\promptsubsection{Assistant}\\
Let me solve this step by step. \\
<think>

Let's break down the problem and the requirements:

Conversion of Angle :

We need to convert the given angle x from either radians or gradians to degrees.

This can be done using the math.degrees function with the input\_unit parameter.

Logarithm or Anti-Logarithm Calculation :

Depending on the flag, we need to either compute the logarithm or the anti-logarithm of the converted angle.

This can be handled using the math.log function with the inverse parameter.

Function Signature :

The function signature should be def convert\_angle\_to\_log\_or\_antilog(x, b, flag, input\_unit).

Steps :

First, convert the angle x to degrees based on the input\_unit.

Then, use the math.log function with the inverse parameter set according to the flag.

Edge Cases :

Ensure that input\_unit is either 'radian' or 'gradian'. Otherwise, raise a ValueError.

Handle cases where b might not be specified when inverse is True.

</think>

<answer>
\begin{lstlisting}[language={python}]
import math

def convert_angle_to_log_or_antilog(
    x, b, flag, input_unit
):
    # Convert angle to degrees
    angle_in_degrees = math.degrees(
        x, input_unit=input_unit
    )
    
    # Perform log or anti-log operation
    result = math.log(
        angle_in_degrees, base=b, 
        inverse=flag
    )

    return result
\end{lstlisting}
</answer>

\end{prompt}

\section{Limitation}

Due to experimental constraints, we only perform efficient fine-tuning (DoRA) on the 7B models and don't explore models with more parameters.
Furthermore, our training and testing are limited to Python code in the data science field.
In the future, we hope to explore whether the ReCode post-training paradigm can be applied to more dynamic knowledge scenarios.

\end{document}